\documentclass[journal]{IEEEtran}
\usepackage{graphicx}
\usepackage{amssymb}%,graphicx}
\usepackage[cmex10]{amsmath}
\usepackage[usenames,dvipsnames]{color}
\usepackage{geometry}                % See geometry.pdf to learn the layout options. There are lots.
\usepackage{epstopdf}
\usepackage{textcomp}
\usepackage{subfig}
\usepackage{url}
\usepackage[bookmarks,colorlinks,breaklinks,linkcolor=Blue,citecolor=Sepia,urlcolor=BlueViolet]{hyperref}
\usepackage{array}
\usepackage{bbold}
\usepackage{hhline}

\long\def\drop#1{}

\let\epsilon\varepsilon

\def\XXint#1#2#3{{\setbox0=\hbox{$#1{#2#3}{\int}$}
     \vcenter{\hbox{$#2#3$}}\kern-.5\wd0}}

\begin{document}

\title{A Regularization Approach to Blind Deblurring and Denoising of QR Barcodes}

\author{Yves~van~Gennip, Prashant~Athavale, J\'er\^ome Gilles, and~Rustum~Choksi
\thanks{Y. van Gennip is with the School of Mathematical Sciences, University of Nottingham, University Park, Nottingham, NG7 2RD, UK, email: y.vangennip@nottingham.ac.uk}
\thanks{P. Athavale is with the Fields Institute, University of Toronto}
\thanks{J. Gilles is with the Department of Mathematics \& Statistics, San Diego State University}
\thanks{R. Choksi is with the Department of Mathematics and Statistics, McGill University}}

%\author{Yves van Gennip\thanks{School of Mathematical Sciences, University of Nottingham, University Park, Nottingham, NG7 2RD, UK, email: y.vangennip@nottingham.ac.uk} \,\,\, \,
%    Prashant Athavale\thanks{Fields Institute, University of Toronto} \,\,\, \,
%	 J\'er\^ome Gilles\thanks{Department of Mathematics \& Statistics, San Diego State University} \,\,\, \,
%         Rustum Choksi\thanks{Department of Mathematics and Statistics, McGill University}}

\markboth{IEEE Transactions on Image Processing ********}%
{A Regularization Approach to Blind Deblurring and Denoising of QR Barcodes}

\maketitle

\begin{abstract}
QR bar codes are prototypical images for which part of the image is a priori known (required patterns). Open source bar code readers, such as {\it ZBar} \cite{ZBar}, are readily available. We exploit both these facts to provide and assess  purely regularization-based methods for blind deblurring of QR bar codes in the presence of noise.
\end{abstract}


%\textbf{Ocis codes: }(070.0070)   Fourier optics and signal processing; (100.0100)   Image processing.

\section{Introduction}
\begin{figure}[h]
\begin{center}
\includegraphics[width=2cm]{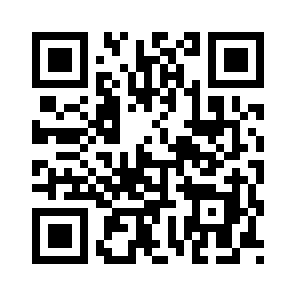}
\caption{A QR bar code (used as ``Code 1'' in the tests described in this paper). Source: Wikipedia \cite{Wiki}}
\label{fig:barcodes0}
\end{center}
\end{figure}

Invented in Japan by the Toyota subsidiary Denso Wave in 1994, QR  bar codes ({\it Quick Response bar codes}) are  a type of matrix 2D bar codes (\cite{Palmer07,QRstuff,Wiki}) that were originally created to track vehicles during the manufacturing process (see Figure~\ref{fig:barcodes0}).
Designed to allow its contents to be decoded at high speed, they have now become the most popular type of matrix 2D bar codes and are easily read by most smartphones.

\begin{figure}[h]
\begin{center}
 \includegraphics[width=0.8\columnwidth]{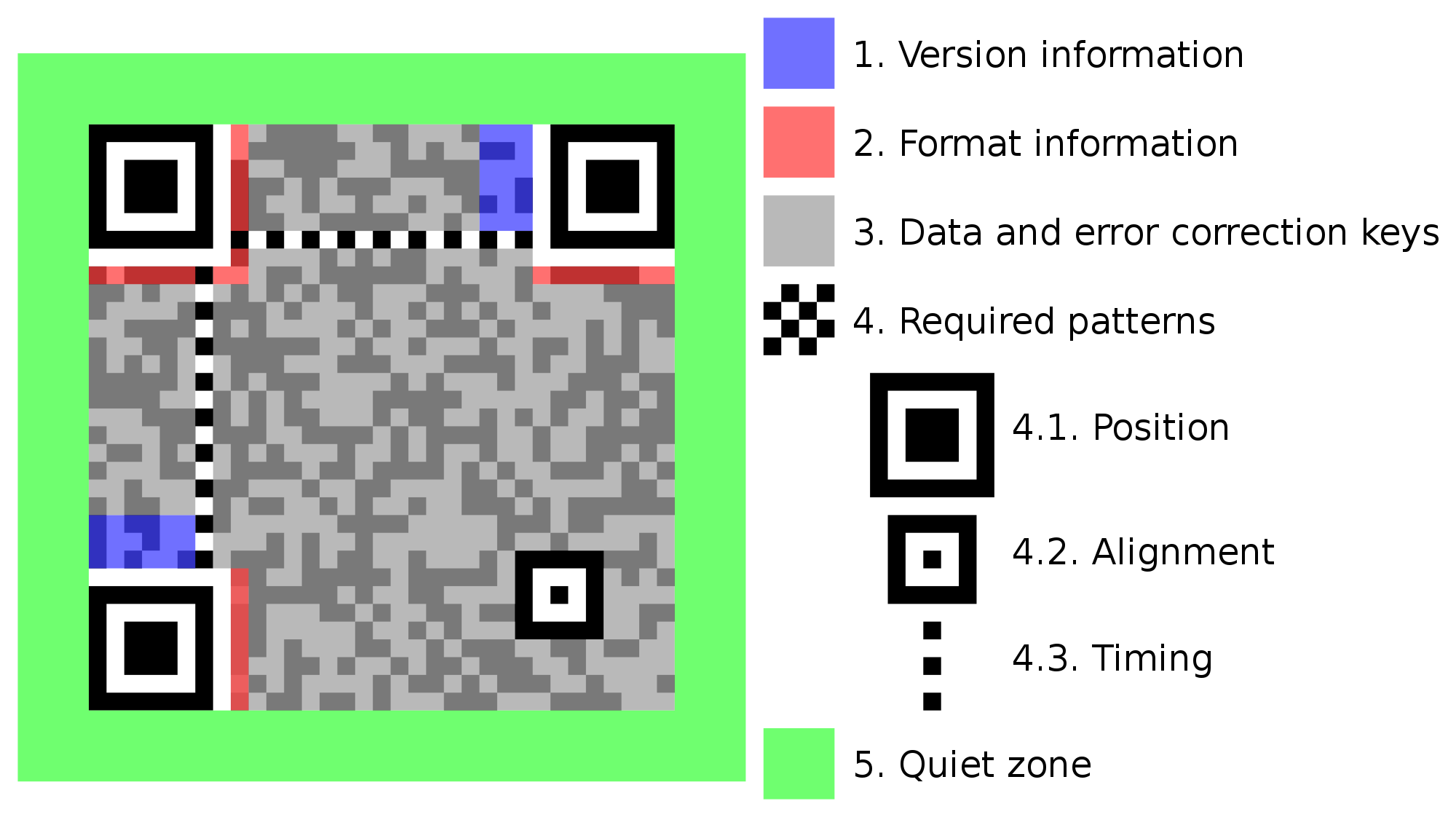}
\caption{Anatomy of a QR bar code. Best viewed in color. Source: Wikipedia \cite{Wiki} (image by Bobmath, CC BY-SA 3.0)}\label{fig:barcodes}
\end{center}
\end{figure}

Whereas standard 1D bar codes are designed to be mechanically scanned by a narrow beam of light, a  QR  bar code is detected as a 2D digital image by a semiconductor image sensor and is then digitally analyzed by a programmed processor (\cite{Palmer07,Wiki}).
Key to this detection are a set of {\it required patterns}. These consist of:
three fixed squares at the top and bottom left corners of the image ({\it finder or position patterns} surrounded by {\it separators}), a smaller square  near the bottom right  corner ({\it alignment pattern}), and  two lines of pixels connecting the two top corners at their bottoms and the two left corners at their right sides ({\it timing patterns}); see
Figure~\ref{fig:barcodes}.

In this article we address blind deblurring and denoising of QR bar codes in the presence of noise. We use the term blind  as our method makes no assumption on the nature, e.g. Gaussian, of the unknown point spread function (PSF) associated with the deblurring.
This is a problem of considerable interest.  While mobile smartphones equipped   with a camera are increasingly used for QR bar code reading,
limitations of the camera imply that the captured images are always blurred and noisy.
A common source of  camera blurring is the relative motion between the camera and bar code. Thus the interplay between deblurring and bar code symbology is important for the successful use of mobile smartphones (\cite{Kato, Eisaku, Thielemann, Turin, ChuYangChen07b,Yang}).

\subsection{Existing  Approaches for  Blind Deblurring of Bar codes}
We note that there are currently a wealth of regularization-based methods for deblurring of general images. For a signal $f$, many attempt to minimize
\[ {\mathcal E} (u, \phi) \, = \, {\mathcal F} (u \ast \phi - f) \, + \,  {\mathcal R}_1 (u) \, + {\mathcal R}_2 (\phi) \]
over all $u$ and PSFs $\phi$, where ${\mathcal F} $ denotes a fidelity term, the ${\mathcal R}_i$ are regularizers, often of total variation (TV) type and $u$ will be the recovered image (cf. \cite{ChanShen05,YouKaveh96,ChanWong98, Molina, Bar, Romero, He, Shan, Takeda}).
Recent work uses sparsity based priors to regularize the images and PSFs (cf. \cite{Bronstein, Tzikas, Cai1, Cai2, Cai3}).

On the other hand, the simple structure of bar codes has lent itself to tailor made deblurring methods, both regularization and non-regularization based.
Much work has been done on 1D bar codes (see for example,  \cite{Joseph, Turin, Kresic1, Kresic2, Kim, Esedoglu04, ChoksiGennip,Iwen}). 2D matrix and stacked  bar codes (\cite{Palmer07})
have received less attention (see for example \cite{Parikh, Xu, Liuetal, BrettElliottDedner14}). The paper of Liu et al \cite{Liuetal} is the closest to the present work,  and proposes an iterative  {\it Increment Constrained Least Squares filter} method for certain 2D matrix bar codes within a Gaussian blurring ansatz. In particular, they use the L-shaped finder pattern of their codes to estimate the standard deviation of the Gaussian PSF, and then restore the image by successively  implementing  a bi-level constraint.

\subsection{Our Approach for QR Bar Codes}
In this article we design a hybrid method, by incorporating the required patterns of the QR symbology into known regularization methods. Given that QR barcodes are widely used in smartphone applications,
we have chosen to focus entirely on these methods because of their simplicity of implementation, modest memory demands, and potential speed.
We apply the method to a large catalogue of corrupted bar codes, and  assess the results with the open source software {\it ZBar} \cite{ZBar}.
%Let $z$ denote the characteristic function of the white area of a QR bar code.
A scan of the bar code  leads to a measured signal $f$, which is a blurry and noisy version of the clean bar code $z$. We assume that $f$ is of the form
\begin{equation}\label{f-def}
f = N(\phi_b*z),
\end{equation} where $\phi_b$ is the PSF (blurring kernel) and $N$ is a noise operator.
Parts of the bar code are assumed known. In particular, we focus on the known top left corner of a QR bar code.  Our goal is to exploit this known information  to accurately estimate the unknown PSF $\phi_b$, and to complement this with state of the art methods  in  TV based regularizations for deconvolution and denoising.
Specifically, we perform the following four steps:
(i)  denoising the signal via a weighted TV flow; (ii) estimating the PSF by a higher-order smooth regularization method based upon comparison of the known finder pattern in the upper left corner with the denoised signal from step (i) in the same corner; (iii) applying appropriately regularized deconvolution with the PSF of step (ii);
(iv) thresholding the output of step (iii).
We also compare the full method with the following subsets/modifications of the four steps: steps (i) and (iv) alone; and the replacement of step (ii) with a simpler estimation based upon a uniform PSF ansatz.

In principle our method extends to blind deblurring and denoising of any class of images
for which a part of the image is {\it a priori known}. We focus on QR bar codes because they present a canonical class of ubiquitous images possessing this property,    their simple binary structure of blocks lends itself well to a simple anisotropic TV regularization, and software  is readily available to both generate and {\it read} QR bar codes, providing  a simple and unambiguous way in which to assess our methods.

\section{TV Regularization and Split Bregman Iteration}\label{sec:background}

Since the seminal paper of Rudin-Osher-Fatemi \cite{RudinOsherFatemi92-2}, TV ({\it i.e.}, the $L^1$ norm of the gradient) based regularization methods have proven to be successful for image
denoising and deconvolution.
Since that time, several improvements have been explored, such as anisotropic (\cite{EsedogluOsher, ChoksiGennipOberman}) and nonlocal
(\cite{GilboaOsher08}) versions of TV. Let us recall the philosophy of such models by describing the anisotropic TV denoising case.

This method constructs a restored image $u_0$ from an observed image $f$ by solving
\begin{equation}\label{eq:tv}
u_0=\underset{u}{\text{argmin}}\|\nabla u\|_1+\frac{\mu}{2}\|u-f\|_2^2,
\end{equation}
where $\|\nabla u\|_1 = |u_x|+ |u_y|$ (here $u_x,u_y$ are the partial derivatives of $u$ in the
$x$ and $y$ directions, respectively).
One of the current state of the art methods for the fast numerical computation of TV based regularization schemes is split Bregman iteration (\cite{Yinetal,Goldstein,Gilles}).
It consists of splitting the above minimization problem into several problems which are easier to solve by introducing extra variables $d_x=u_x$ and $d_y=u_y$ (the subscripts in the new variables do not denote differentiation). Solving for \eqref{eq:tv} is equivalent
to finding $u_0,d_x,d_y$ which solve the following steps (\cite{Yinetal,Goldstein}):
\begin{enumerate}
\item $(u^{k+1},d_x^{k+1},d_y^{k+1})= \text{argmin} \|d_x\|_1+\|d_y\|_1$ $+\frac{\mu}{2}\|u-f\|_2^2
+\frac{\lambda}{2}\|d_x-u_x+b_x^k\|_2^2+\frac{\lambda}{2}\|d_y-u_y+b_y^k\|_2^2$
\item $b_x^{k+1}=b_x^k+u_x^{k+1}-d_x^{k+1}$
\item $b_y^{k+1}=b_y^k+u_y^{k+1}-d_y^{k+1}$
\end{enumerate}
Then $u_0$ will be given by the final $u^k$.

The first step can be handled by alternatively fixing two of the variables and then solving for the third one. It is easy to see that if we fix $d_x$ and $d_y$ we get an $L^2-L^2$ problem which can be efficiently solved
by a Gauss-Seidel scheme \cite{Goldstein} or in the Fourier domain \cite{Gilles}. Here we take  the Fourier approach.
 The updates $d_x^{k+1},d_y^{k+1}$ correspond to the solution of $L^1-L^2$ type problems and are given by
\begin{gather*}
d_x^{k+1}=shrink(u_x^{k+1}+b_x^k,1/\lambda)\\
d_y^{k+1}=shrink(u_y^{k+1}+b_y^k,1/\lambda)
\end{gather*}
where the operator $shrink$ is given by $shrink(v,\delta)=sign(v)\max(0,|v|-\delta)$.

\section{The Method}\label{sec:method}

\subsection{Creating Blurred and Noisy QR Test Codes}
We ran our tests on a collection of QR bar codes, some of which are shown in Figure~\ref{fig:morecleancodes}. In each case, the clean bar code is denoted by $z$. The
{\it module width} denotes the
 length of the smallest square  of the bar code (the analogue of  the $X$-dimension in a 1D bar code). In each of the QR codes we used for this paper,
 %for example those in Figure~\ref{fig:morecleancodes},
 this length consists of 8 pixels. In fact, we can automatically extract this length from the clean corners or the timing lines in the bar codes.

We create blurry and noisy versions of the clean bar code $z$ as follows. We use MATLAB's function ``fspecial'' to create the blurring kernel $\phi_b$. In this paper we discuss results using
%Gaussian blur (a rotationally symmetric Gaussian lowpass filter with prescribed size and standard deviation)
isotropic Gaussian blur (with prescribed size and standard deviation) and motion blur (a filter which approximates, once convolved with an image, the linear motion of a camera by a prescribed number of pixels, with an angle of thirty degrees in a counterclockwise direction). The convolution is performed using MATLAB's ``imfilter($\cdot$,$\cdot$,`conv')'' function.

In our experiments we apply one of four types of noise operator $N$ to the blurred bar code $\phi_b*z$:
\begin{itemize}
\item {\it Gaussian noise} (with zero mean and prescribed standard variation) via the addition to each pixel of the standard variation times a pseudorandom number drawn from the standard normal distribution (MATLAB's function ``randn'');
\item {\it uniform noise} (drawn from a prescribed interval), created by adding a uniformly distributed pseudorandom number (via MATLAB's function ``rand'') to each pixel;
\item  {\it salt and pepper noise} (with prescribed density), implemented using MATLAB's ``imnoise'' function;
 \item {\it speckle noise} (with prescribed variance), implemented using MATLAB's ``imnoise'' function.
\end{itemize}
Details for the  blurring and noise parameters  used in our tests  are given in Section~\ref{sec:results}.

We denote the region of the finder pattern in the upper left corner of a barcode by $C_1$\footnote{To be precise, $C_1$ is the square region formed by the upper left finder patter, its separator, and the part of the quiet zone needed to complete $C_1$ into a square. For simplicity we will refer to the pattern in $C_1$ as a finder pattern.} and note that the part of the (clean) bar code which lies in this region is known a priori.
To denoise and deblur, we now apply the following 4-step process to the signal $f$ defined by (\ref{f-def}).

 \subsection{Step (i): Denoising via weighted TV flow}\label{sec:denoisingviaTVflow}

Preliminary experiments showed that if we perform denoising and kernel estimation at the same time, the results are much worse than when we dedicate separate steps to each procedure. Hence, we first denoise the image using the weighted TV flow\footnote{In earlier stages of experimentation we used a nonlocal TV approach for denoising (code from \cite{Mao2012c}, based on the split Bregman technique described in \cite{Zhang}). Due to the edge-preserving property,  the weighted TV flow technique described above gives better results with less rounded corners and so it is the one we have used to produce the results in this paper.}
(\cite{athavale, XA, athavaleMIA}),
\begin{equation}\label{WTV}
\frac{\partial u}{\partial t}=\mu(t) \mathrm{div}\Big( \frac{\alpha \nabla u }{|\nabla u|}\Big)
\end{equation}
with Neumann boundary conditions. Here the initial condition for $u(x,t)$ is given by $u(x,0) = f$ and $\alpha$ is a diffusion controlling function,
$$ \alpha(x):=\frac{1}{\sqrt{1+\frac{|(G_{\sigma}*\nabla f)(x)|^2}{\beta^2}}},$$ where $G_\sigma$ is a normalized Gaussian function with mean zero and standard deviation $\sigma$.

The flow \eqref{WTV} is closely related to the standard total variation flow (\cite{andreu01, ACDM}), and can be obtained \cite{athavale} as a limiting case of hierarchical $(BV_{\alpha}, L^2)$ decomposition of $f$. Here, the $BV_\alpha$ semi-norm of $u$ is defined as $|u|_{BV_{\alpha}}:=\int_{\Omega}\alpha |\nabla u|$. In the hierarchical $(BV_{\alpha}, L^2)$ decomposition of $f$, finer scale components are removed from $f$ successively.
\begin{figure}[ht]
\begin{center}
\includegraphics[trim={3cm 8.3cm 3.7cm 8.7cm}, clip, width=0.4\textwidth]{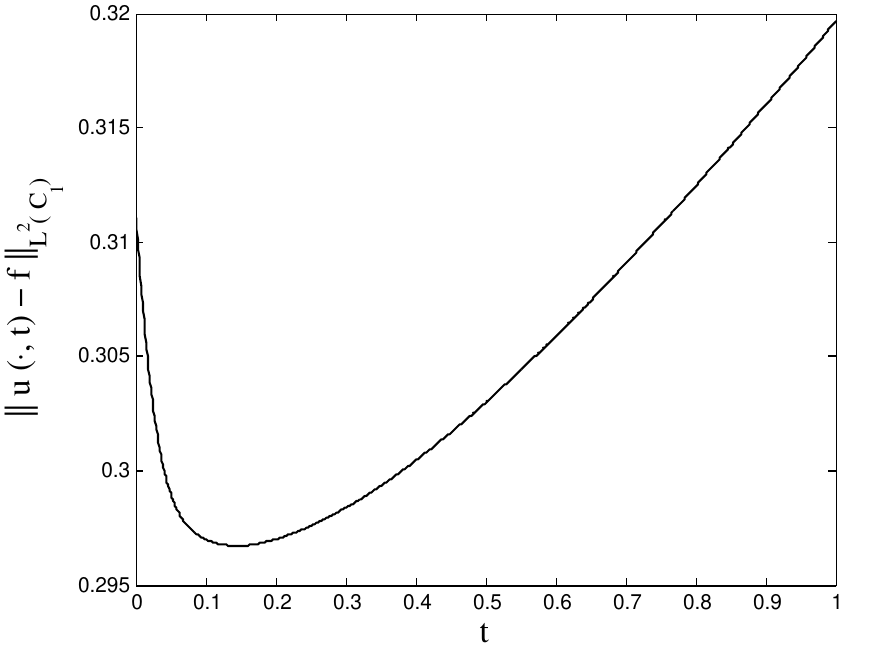}
\caption{A typical run of the weighted $TV$ flow, over which $\| u(\cdot, t) - f \|_{L^2(C_1)}$ first decreases and then increases again.}
\label{fig:TVsim}
\end{center}
\end{figure}

The weight function $\alpha$ reduces the diffusion at prominent edges of the given QR image, i.e. at points $x$ where the gradient $|\nabla f|$ is high. The Gaussian smoothing avoids false characterization of noise as edges, and the value of $\beta$ is chosen so that $\alpha(x)$ is small at relevant edges in the image. The function $\mu(t)$ in \eqref{WTV} is the monotonically increasing ``speed'' function. It can be shown (cf. \cite{athavale}) that the speed of the flow is directly dependent on $\mu(t)$; more precisely $\Vert\frac{\partial u}{\partial t}\Vert_{\alpha^*}=\mu(t)$, where $\Vert\cdot\Vert_{\alpha^*}$ denotes the dual of the weighted $BV_{\alpha}$ semi-norm, with respect to the $L^2$ inner product.

We use homogeneous Neumann boundary conditions on the edges of the image (following \cite{TadmorNezzarVese04,TadmorNezzarVese08}), which are implemented in practice by extending the image outside its boundary with the same pixel values as those on the boundary.

%The flow $u(\cdot, t)$ reaches a constant value $f_{avg}$ in finite time, $T_f$ (cf. \cite{ACDM}).
Since $u(., t)$ is an increasingly smooth version of $f$ as t increases, we need to decide the stopping time $t_s$ for which $u(\cdot, t_s)$ is a (near) optimally smoothed version of $f$. We take advantage of the fact that the upper left corner $C_1$ is known a priori. We stop the flow at time $t_s$ when $\| u(\cdot, t) - f \|_{L^2(C_1)}$ attains its minimum, and the function $u_1 := u(\cdot, t_s)$ is taken as a denoised version of the QR code for further processing.  This phenomenon is illustrated in Fig. \ref{fig:TVsim}. Thus, one of the advantages of using the weighted TV flow, over other minimization techniques, is that we do not need to change the denoising parameters to get optimal results.

For the experiments in this paper, we choose $\beta=0.05$, which corresponds to reduced regularization
at relevant edges in the image $f$. The standard deviation of the Gaussian is empirically set to $\sigma=1$. The weighted TV flow successively removes small-scale structures from the original image $f$, as $t$ increases.
As noise is usually the finest-scale structure in an image, we expect that the weighted TV flow removes the noise for smaller values of $t$ (see Figure \ref{fig:TVsim}). In order to effectively remove small-scale noise, while avoiding over-smoothing, we need to control the magnitude of speed function $\mu(t)$ for small values of $t$. Therefore, we choose $\mu(t)=1.1^t$, making the speed $\Vert\frac{\partial u}{\partial t}\Vert_{\alpha^*} \approx 1$ for small values of $t$.

\subsection{Step (ii):  PSF estimation}\label{sec:PSFestimation}

To determine the blurring kernel we compare the known finder pattern in the upper left corner $C_1$ of $z$ with the same corner of $u_1$:
\[
\phi_* = \underset{\phi}{\text{argmin}}\, \frac12 \int_{C_1} |\nabla \phi|^2 + \frac{\lambda_1}2 \|\phi*z-u_1\|_{L^2(C_1)}^2.
\]
Here $\lambda_1>0$ is a parameter whose choice we will discuss in more detail in Sections~\ref{sec:paramest} and~\ref{sec:varyinglambda1}.

Note that this variational problem is strictly convex and hence possesses a unique minimizer which solves the Euler-Lagrange equation
\[
-\Delta\phi_*+\lambda_1(\phi_*\ast z-u_1)\ast z=0.
\]
Solving the equation in Fourier space, with the hat symbol denoting the Fourier transform, we find
\begin{gather*}
-\hat{\Delta}\widehat{\phi_*}+\lambda_1(\widehat{\phi_*}\hat{z}-\hat{u_1})\hat{z}=0\\
\Leftrightarrow (-\hat{\Delta}+\lambda_1|\hat{z}|^2)\widehat{\phi_*}-\lambda_1\hat{u_1}\hat{z}=0\\
\Leftrightarrow \widehat{\phi_*}=(-\hat{\Delta}+\lambda_1|\hat{z}|^2)^{-1}\lambda_1\hat{u_1}\hat{z}.
\end{gather*}
We take the inverse Fourier transform to retrieve $\phi_*$.
To ensure that the kernel is properly normalized, i.e. $\int \phi_* = 1$, we normalize the input signals $u_1$ and $z$ to unit mass before applying the algorithm.

To reduce the influence of boundary effects in the above procedure, we impose periodic boundary conditions, by extending (the corners of) both $z$ and $u_1$ with their mirror images on all sides, and extracting the center image, corresponding to the original image (corner) size, from the result after minimization.

The output of the algorithm is of the same size as $C_1$, even though the actual kernel is smaller. To make up for this we reduce the size of the output kernel by manual truncation  (i.e., setting all entries, except those in the center square of specified size, in the kernel matrix to zero). We determine this smaller size automatically by constructing a collection $\{\phi_1, \phi_2\, \ldots, \phi_n\}$ of kernels of different sizes by reducing $\phi_*$ and selecting the $\phi_i$ for which $\|\phi_i*z - u_1\|_{L^2(C_1)}^2$ is minimized.  This comparison is fairly quick and in practice we compare kernels of many or all sizes up to the full size of the original output. It is important to normalize the reduced kernel again to unit mass. This normalized kernel is denoted by $\phi_*$ in the sequel.

\subsection{Step (iii): Deblurring (Deconvolution)}\label{sec:deblurringstep}

With the minimizing kernel $\phi_*$, we then proceed to deblur the denoised signal $u_1$ via the variational problem
\[
u_2 = \underset{u}{\text{argmin}}\, \int |u_x| + |u_y| + \frac{\lambda_2}2 \|\phi_**u-u_1\|_{L^2}^2.
\]
We will discuss the choice of the parameter $\lambda_2>0$ in Section~\ref{sec:paramest}.

This minimization problem is solved via the split Bregman iteration method \cite[ATV\_NB\_Deconvolution]{Gilles}, described previously in Section~\ref{sec:background}, which uses an internal fidelity parameter for the split variables, whose value we set to 100. We use two internal Bregman iterations.
We again impose periodic boundary conditions, by extending $u$ and $u_1$ with mirror images, and extracting the center image, corresponding to the original image size, from the result.

\subsection{Step (iv): Thresholding}\label{sec:thresholdingstep}

\begin{figure}
\begin{center}
\subfloat[The unprocessed bar code]{
\includegraphics[width=0.145\textwidth]{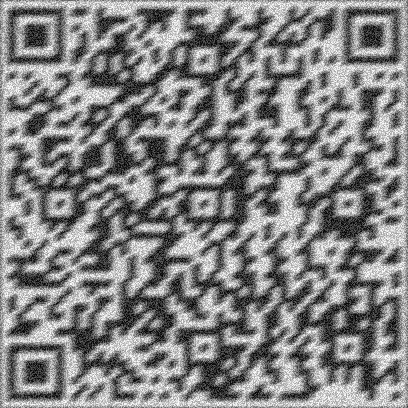}}\hspace{0.05cm}
\subfloat[The cleaned bar code, thresholded per block]{
\includegraphics[width=0.145\textwidth]{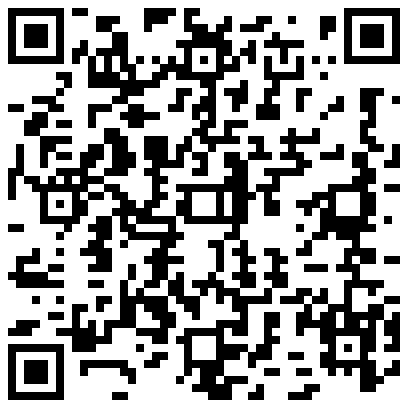}}\hspace{0.05cm}
\subfloat[The cleaned bar code, thresholded per pixel]{
\includegraphics[width=0.145\textwidth]{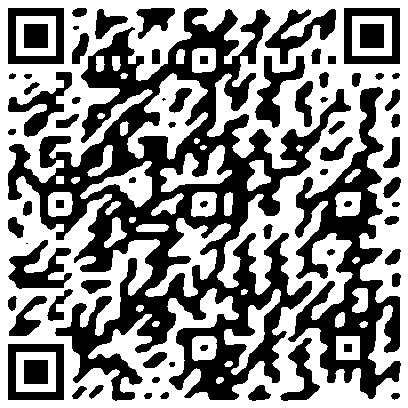}}
\end{center}
\caption{Code 10 with motion blur (length 11) and uniform noise (sampled from $[0,0.6]$). Neither the blurry and noisy bar code (a), nor the cleaned code that is thresholded per block (b), can be read by {\it ZBar}. The cleaned code that is thresholded per pixel (c) can be read.}
\label{fig:thresholdingcomparison}
\end{figure}
Finally we threshold the result based on the gray value per pixel to get a binary image $u_3$. We determine the threshold value automatically, again using the a priori knowledge of $z|_{C_1}$, by trying out a set of different values (100 equally spaced values varying from the minimum to the maximum gray value in $u_3|_{C_1}$) and picking the one that minimizes $\|u_3-z\|_{L^2(C_1)}^2$.

Instead of thresholding per pixel, we also experimented with thresholding the end result based on the average gray value per block of pixels of size module width by module width, again using an empirically optimized threshold value. Our tests with Zbar show significant preference for thresholding per pixel rather than per block, despite the output of the latter being more `bar code like' (cf. Figure~\ref{fig:thresholdingcomparison}). In the remainder of this paper we will only consider thresholding per pixel.
In the final output we add back the known parts of the QR bar code, i.e., the required patterns of Figure~\ref{fig:barcodes}.

\subsection{Fidelity parameter estimation}\label{sec:paramest}

We have introduced two fidelity parameters $\lambda_1$ and $\lambda_2$. In the original Rudin-Osher-Fatemi model \cite{RudinOsherFatemi92-2}, if the data is noisy, but not blurred, the fidelity parameter $\lambda$ should be inversely proportional to the variance of the noise \cite[Section 4.5.4]{ChanShen05}. Hence we set $\lambda_2 = 1/\sigma_2^2$, where $\sigma_2^2$ is the variance of $(\phi_**z-u_1)\big|_{C_1}$.
The heuristic for choosing the fidelity parameter $\lambda_2$ is based on the assumption that the signal $u_1$ is not blurred. Hence the expectation, which was borne out by some testing, is that this automatic choice of $\lambda_2$ works better for small blurring kernels. For larger blurring we could possibly improve the results by manually choosing $\lambda_2$ (by trial and error), but the results reported in this paper are all for automatically chosen $\lambda_2$.
Very recently we became aware of an, as yet unpublished, new statistical method \cite{Mead15} which could potentially be used to improve on our initial guess for $\lambda_2$ in future work.

Because we do not know the `clean' true kernel $\phi_b$,
we cannot use the same heuristics to choose $\lambda_1$.
Based on some initial trial and error runs, we use $\lambda_1=10000$ in our experiments unless otherwise specified. In Section~\ref{sec:varyinglambda1} we address the question how the performance of our algorithm varies with different choices of $\lambda_1$.

\section{Results}\label{sec:results}

We applied our algorithm to a class of blurred and noisy bar codes with either Gaussian blur or motion blur, and four types of noise (Gaussian, uniform, speckle, and salt and pepper). We then tested the results using Zbar, comparing readability for the original and cleaned signals.

\subsection{Choice of noise and blurring parameters}

Our original clean binary images are normalized to take values in $\{0,1\}$ ($\{\text{black, white}\})$.
To construct a rotationally symmetric, normalized, Gaussian blur kernel with MATLAB's function ``fspecial'', we need to specify two free parameters: the size (i.e. the number of rows) of the square blurring kernel matrix and the standard deviation of the Gaussian. We vary the former over the values in $\{3, 7, 11, 15, 19\}$, and the latter over the values in $\{3, 7, 11\}$.

To construct a motion blur matrix with ``fspecial'', we need to specify the angle of the motion and the length of the linear motion. For all our experiments we fix the former at 30 degrees and vary the latter over $\{3, 7, 11, 15, 19\}$ pixels.

Because our Gaussian blur kernels have one more free parameter than our motion blur kernels, we use a larger range of noise parameters for the latter case.

To create Gaussian noise with zero mean and standard deviation $s$, we add $s$ times a matrix of i.i.d. pseudorandom numbers (drawn from the standard normal distribution) to the image. For Gaussian blur we use $s=0.05$, while for the motion blur  $s$ varies over $\{0, 0.05, 0.1, \ldots, 0.5\}$.

To construct uniform noise, we add $a$ times a matrix of uniformly distributed pseudorandom numbers to the image. This generates uniformly distributed additive noise drawn from the interval $[0,a]$. In the Gaussian blur case, we run tests with $a\in \{0.4, 0.7, 1\}$. In the motion blur case we vary $a$ over $\{0, 0.2, 0.4, \ldots, 2\}$.

Salt and pepper noise requires the input of one parameter: its density. In the Gaussian blur case, we fix the density at 0.2; in the motion blur case we vary it over $\{0, 0.1, 0.2, \ldots,  1\}$.

To produce speckle noise we specify its variance. In the Gaussian blur case, we fix this at 0.4; in the motion blur case we vary the variance over $\{0, 0.1, 0.2, \ldots,  1\}$.

Each of the resulting blurry and noisy bar codes is again normalized so that each pixel has a gray value between 0 and 1.

\subsection{Some examples of cleaned bar codes}

\begin{figure}
\begin{center}
\subfloat[]{\label{fig:GbAn}
\includegraphics[width=0.23\columnwidth]{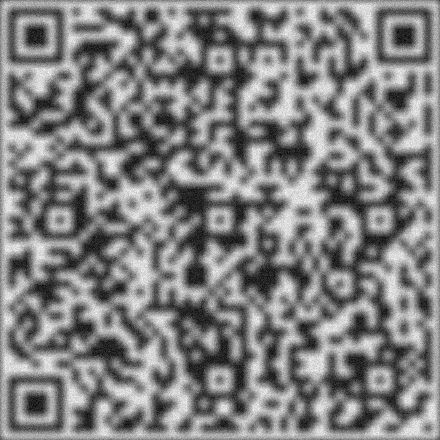}}
\subfloat[]{
\includegraphics[width=0.23\columnwidth]{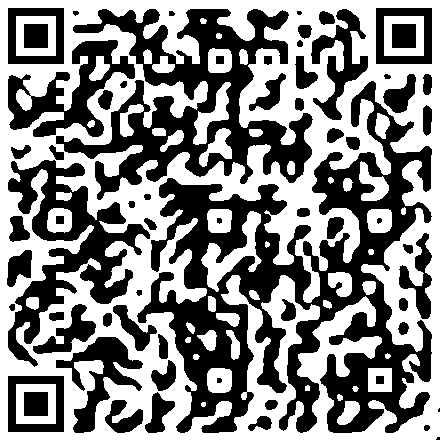}}
\subfloat[]{\label{fig:GbGn}
\includegraphics[width=0.23\columnwidth]{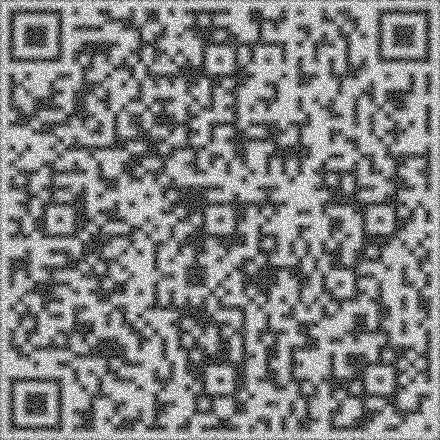}}
\subfloat[]{
\includegraphics[width=0.23\columnwidth]{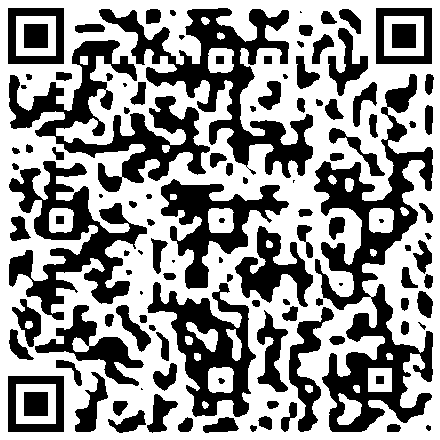}}\\
\subfloat[]{\label{fig:GbSPn}
\includegraphics[width=0.23\columnwidth]{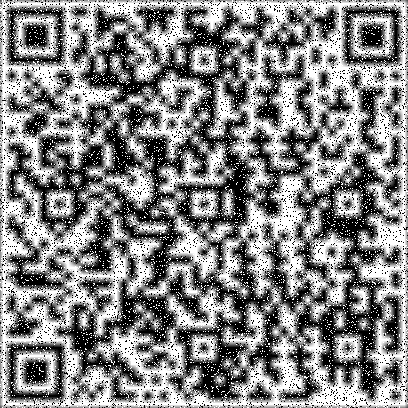}}
\subfloat[]{
\includegraphics[width=0.23\columnwidth]{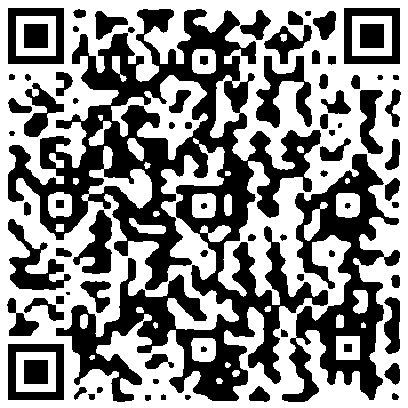}}
\subfloat[]{\label{fig:GnSn}
\includegraphics[width=0.23\columnwidth]{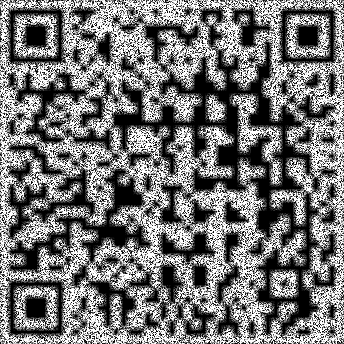}}
\subfloat[]{
\includegraphics[width=0.23\columnwidth]{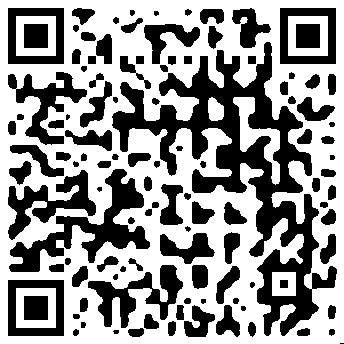}}

\end{center}
\caption{Examples  with Gaussian blur and various types of noise. {\bf(a)-(b):} unprocessed and cleaned Code 7 with Gaussian blur (size 11, standard deviation 11) and Gaussian noise (mean 0, standard variation 0.05), {\bf(c)-(d):} unprocessed and cleaned Code 7 with Gaussian blur (size 7, standard deviation 3) and uniform noise (sampled from $[0, 0.7]$), {\bf(e)-(f):} unprocessed and cleaned Code 10 with Gaussian blur (size 7, standard deviation 11) and salt and pepper noise (density 0.2), {\bf(g)-(h):} unprocessed and cleaned Code 5 with Gaussian blur (size 7, standard deviation 7) and speckle noise (variance 0.6)}
\label{fig:Gaussianblur}
\end{figure}

\begin{figure}
\begin{center}

\subfloat[]{\label{fig:MbAn}
\includegraphics[width=0.23\columnwidth]{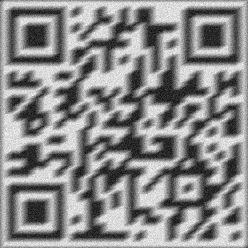}}
\subfloat[]{
\includegraphics[width=0.23\columnwidth]{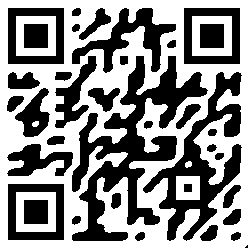}}
\subfloat[]{\label{fig:MbGn}
\includegraphics[width=0.23\columnwidth]{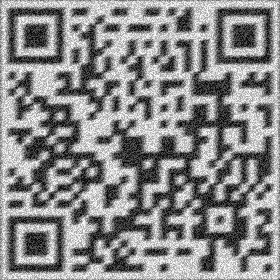}}
\subfloat[]{
\includegraphics[width=0.23\columnwidth]{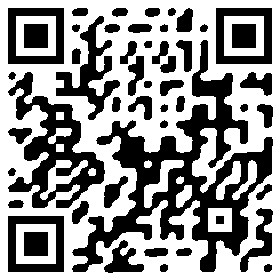}}\\
\subfloat[]{\label{fig:MbSPn}
\includegraphics[width=0.23\columnwidth]{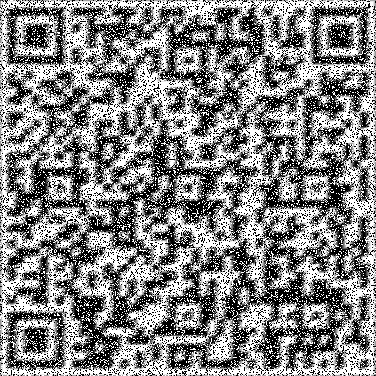}}
\subfloat[]{
\includegraphics[width=0.23\columnwidth]{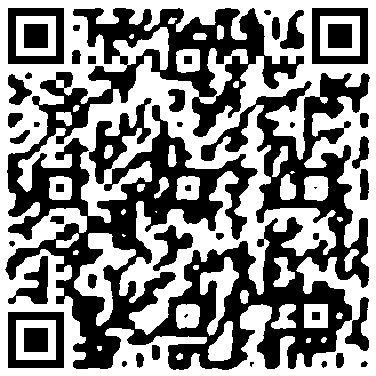}}
\subfloat[]{\label{fig:MbSn}
\includegraphics[width=0.23\columnwidth]{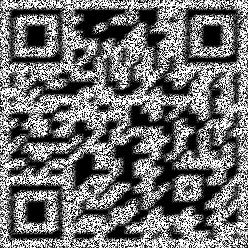}}
\subfloat[]{
\includegraphics[width=0.23\columnwidth]{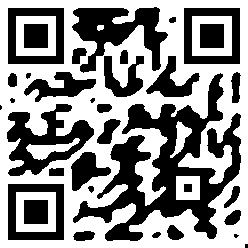}}

\end{center}
\caption{Examples  with motion blur and various types of noise. {\bf(a)-(b):} unprocessed and cleaned Code 2 with motion blur (length 11, angle 30) and Gaussian noise (mean 0, standard deviation 0.05), {\bf(c)-(d):} unprocessed and cleaned Code 3 with motion blur (length 7, angle 30) and uniform noise (sampled from $[0, 0.6]$), {\bf(e)-(f):} unprocessed and cleaned Code 4 with motion blur (length 7, angle 30) and salt and pepper noise (density 0.4), {\bf(g)-(h):} unprocessed and cleaned Code 9 with motion blur (length 7, angle 30) and speckle noise (variance 0.6)}
\label{fig:Motionblur}
\end{figure}

 Figure~\ref{fig:Gaussianblur} shows QR bar codes with Gaussian blur and various types of noise, together with their cleaned output of our algorithm. In all these cases the software {\it ZBar} was not able to read the blurred and noisy bar code, but was able to read the cleaned output. Figure~\ref{fig:Motionblur} shows similar examples for QR bar codes with motion blur. We  stress that we do not use subjective aesthetic pleasantness as the arbiter of whether our reconstruction is good or not, but whether or not the output can be read by {\it ZBar}.

\subsection{Results of our algorithm}\label{sec:resultsalg}

\begin{figure}
\begin{center}

\subfloat[``Code 2'']{\label{fig:code2}
\includegraphics[width=0.25\columnwidth]{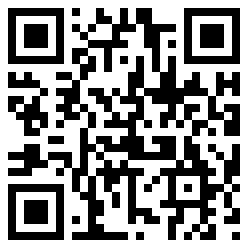}}
\subfloat[``Code 9'']{\label{fig:code9}
\includegraphics[width=0.25\columnwidth]{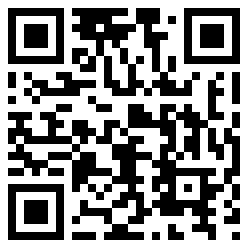}}
\subfloat[``Code 10'']{\label{fig:code10}
\includegraphics[width=0.25\columnwidth]{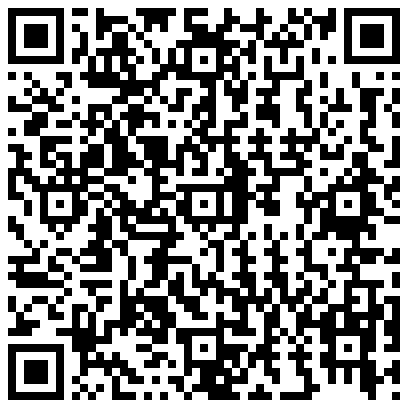}}

\end{center}
\caption{Some of the clean bar codes used in the experiments}
\label{fig:morecleancodes}
\end{figure}

To check that our results are not heavily influenced by the specifics of the original clean bar code or by the particular realization of the stochastic noise, we run our experiments on ten different bar codes (three are shown in Figure~\ref{fig:morecleancodes}), per code averaging the results (0 for ``not readable by {\it ZBar}'' and 1 for ``readable by {\it ZBar}'') over ten realizations of the same stochastic noise. These tests lead to the sixteen `phase diagrams' in Figures~\ref{fig:diaggaussadd}--\ref{fig:diagmotionspeckle}. Note that each diagram shows the average results over ten runs for one particular choice of bar code. Dark red indicates that the {\it ZBar} software is able to read the bar code (either the blurry/noisy one or the cleaned output of our algorithm) in all ten runs, and dark blue indicates that {\it ZBar} cannot read any of the ten runs\footnote{Note that the bar codes displayed in Figures~\ref{fig:Gaussianblur} and~\ref{fig:Motionblur} are chosen from parts of the phase diagrams where the output is readable in all runs.}. While each phase diagram is derived using only one of the test bar codes, our results showed robustness with respect to the  choice.

Before we delve into the specific outcomes, note one general trend for motion blur: in the absence of noise and at high blurring (length 15), the cleaned bar codes are not readable, while the uncleaned ones are. Running our algorithm on the noiseless images without including denoising Step (i) (Section~\ref{sec:denoisingviaTVflow}) does not fix this issue. In the case of Gaussian blur this issue does typically not occur. We return to this issue and its remedy in Section~\ref{sec:3methods}.
In the presence of even the slightest noise this issue disappears. Noise generally renders the uncleaned bar codes completely unreadable (with a minor exception for low level Gaussian noise in Figure~\ref{fig:diagmotionadd}). Those are the situations in which our algorithm delivers.

\subsubsection{Gaussian noise}

\begin{figure}[h]
\begin{center}

\subfloat{\label{fig:diaggaussaddunprocessed}
\includegraphics[trim={0.5cm 6.9cm 0.7cm 6.8cm}, clip, width=0.48\columnwidth]{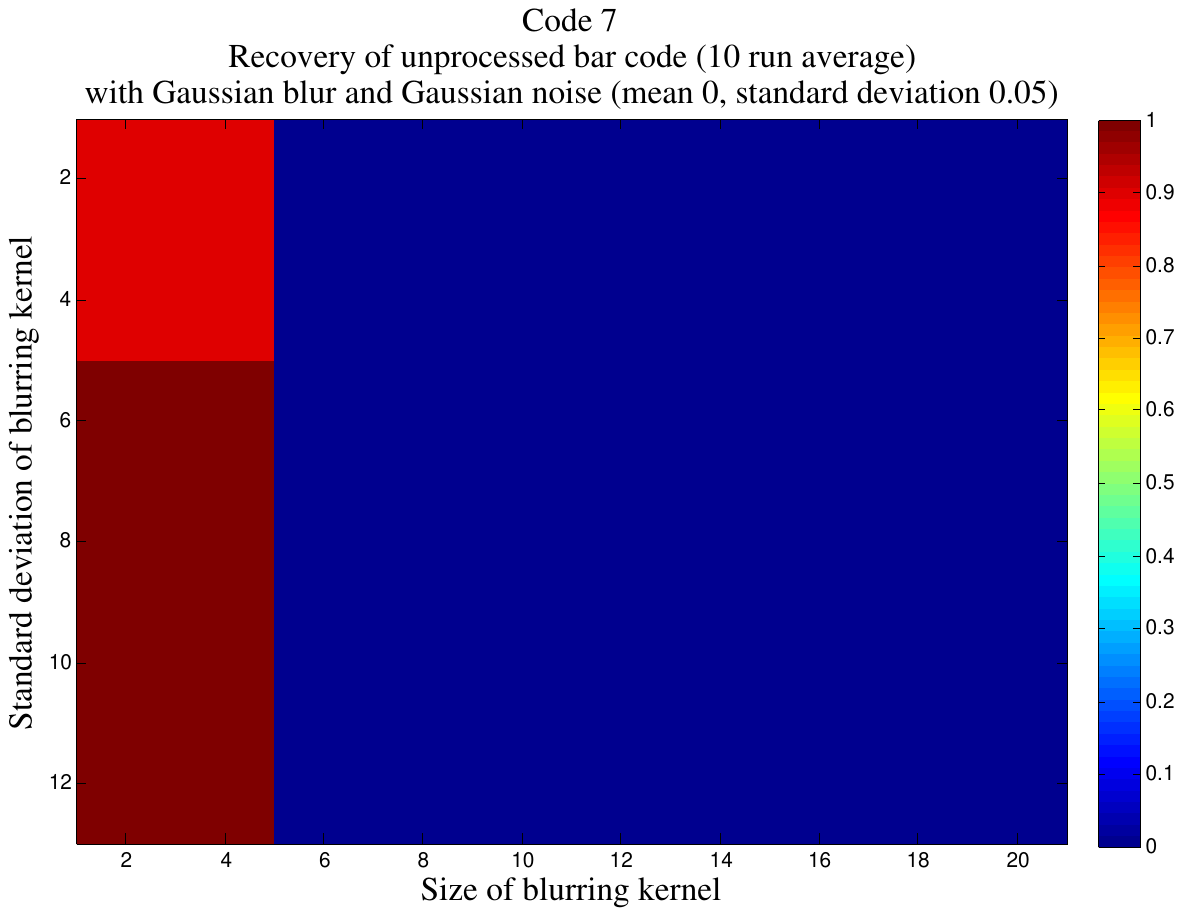}}
\subfloat{\label{fig:diaggaussaddcleaned}
\includegraphics[trim={0.5cm 6.9cm 0.7cm 6.8cm}, clip, width=0.48\columnwidth]{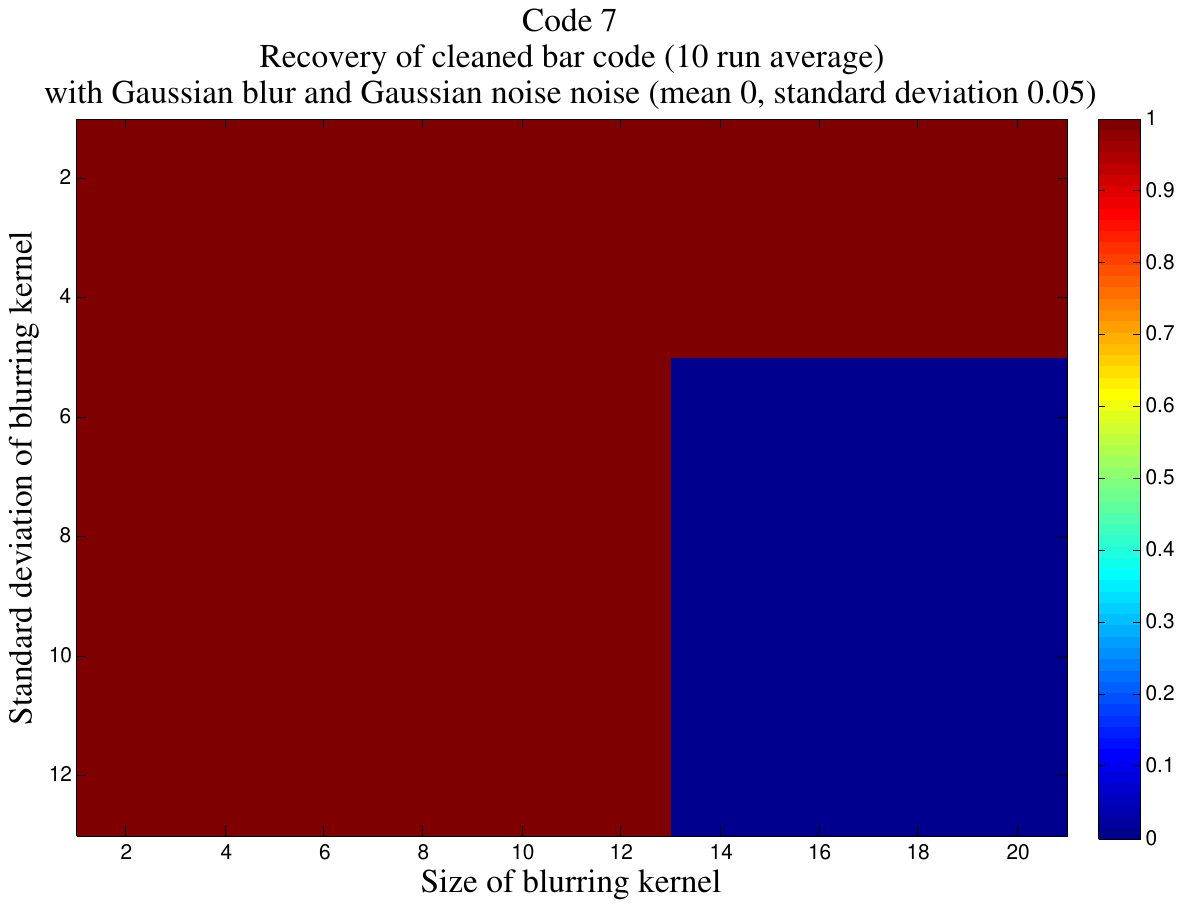}}

\end{center}
\caption{Readability of the unprocessed and cleaned Code 7, for various sizes and standard deviations of the Gaussian blur kernel, with Gaussian noise (mean 0 and standard deviation 0.05). The color scale indicates the fraction of ten runs with different noise realizations which lead to a bar code readable by {\it ZBar}.}
\label{fig:diaggaussadd}
\end{figure}

From Figure~\ref{fig:diaggaussadd}, we see that {\it ZBar} performs very badly in the presence of Gaussian noise even at a low standard deviation of 0.05 for the noise. The bar code is only readable for a small blurring kernel. The cleaned codes however are consistently readable for quite a large range of blurring kernel size and standard deviation.

\begin{figure}[h]
\begin{center}

\subfloat{\label{fig:diagmotionaddunprocessed}
\includegraphics[trim={0.2cm 6.9cm 0.7cm 6.8cm}, clip, width=0.48\columnwidth]{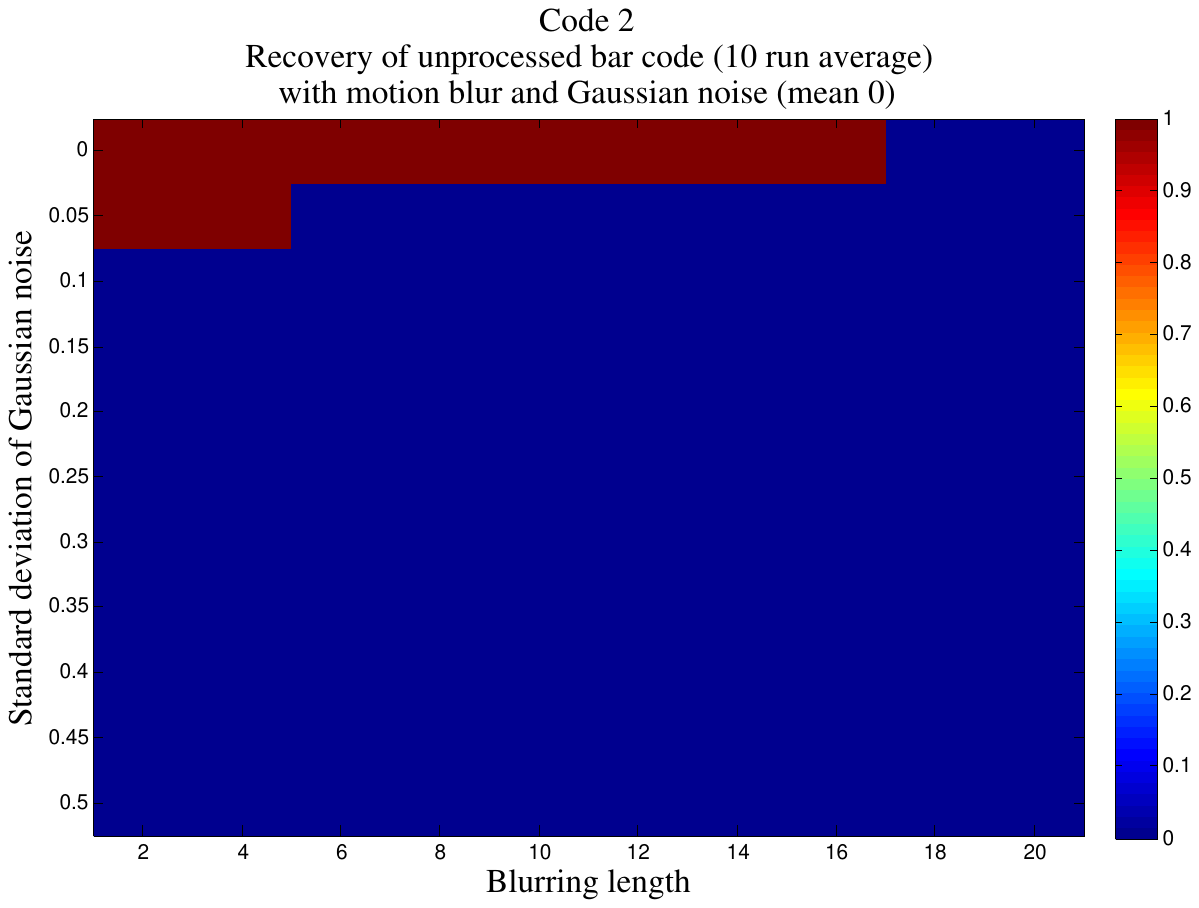}}
\subfloat{\label{fig:diagmotionaddcleaned}
\includegraphics[trim={0.2cm 6.9cm 0.7cm 6.8cm}, clip, width=0.48\columnwidth]{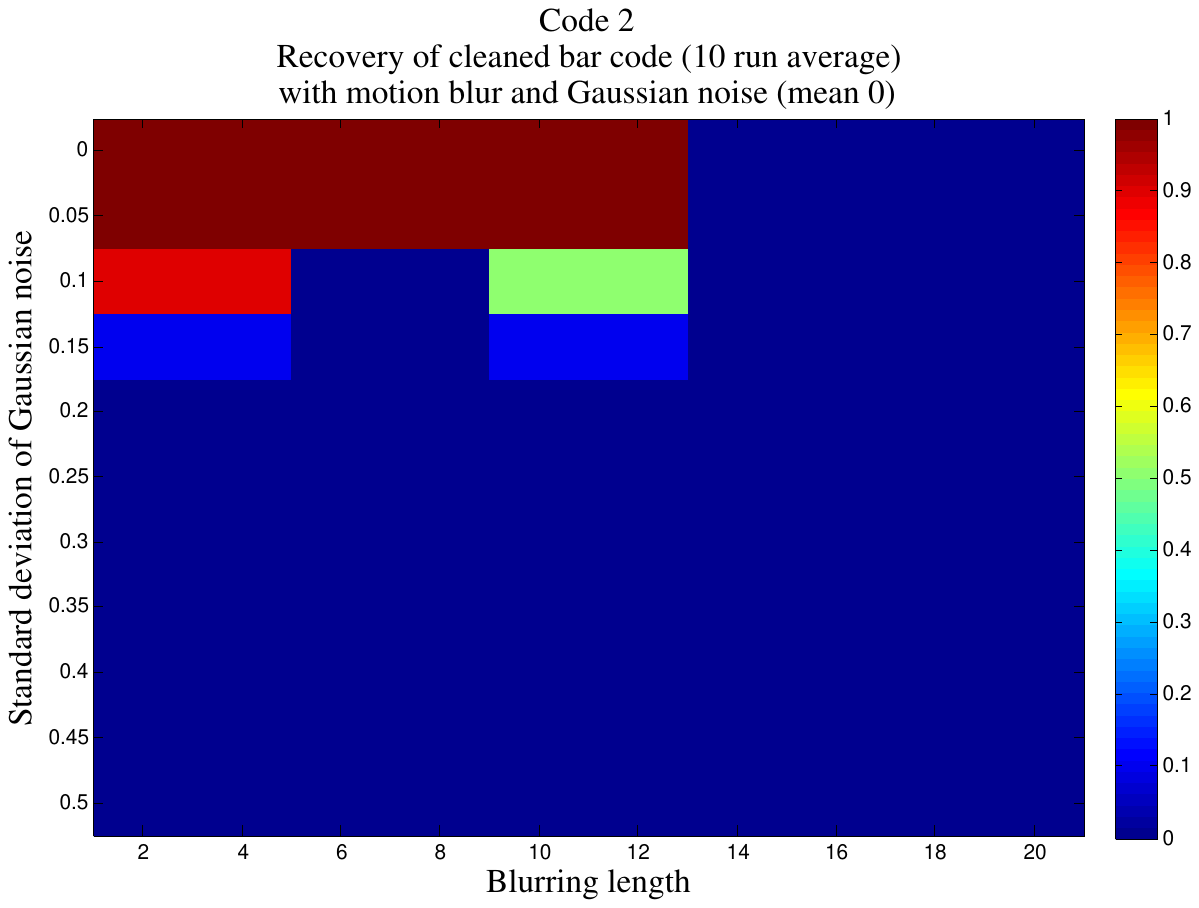}}

\end{center}
\caption{Readability for Code 2 from Figure~\ref{fig:code2}, for various blurring lengths and various values of the Gaussian noise's standard deviation (its mean is 0)}
\label{fig:diagmotionadd}
\end{figure}

For motion blur (cf. Figure~\ref{fig:diagmotionadd}) there is a slight, but noticeable improvement in readability for noise with low standard deviation at larger size blur kernels. Comparing with the motion blur results below, Gaussian noise is clearly the type with which our algorithm has the  most difficulties.

\subsubsection{Uniform noise}\label{sec:uniformnoise}

\begin{figure}[h]
\begin{center}

\subfloat{\label{fig:diaggaussgaussunprocessed}
\includegraphics[trim={0.5cm 6.9cm 0.7cm 6.8cm}, clip, width=0.48\columnwidth]{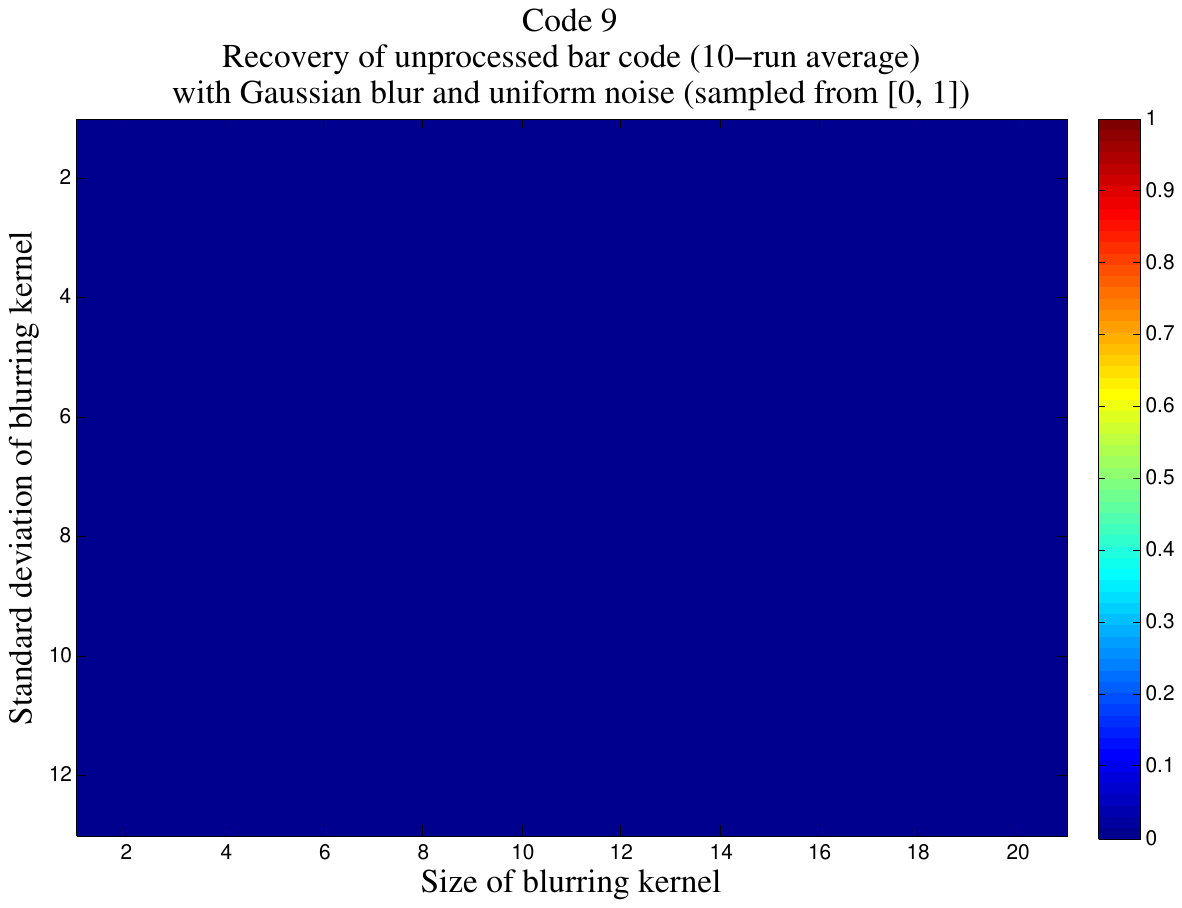}}
\subfloat{\label{fig:diaggaussgausscleaned}
\includegraphics[trim={0.5cm 6.9cm 0.7cm 6.8cm}, clip, width=0.48\columnwidth]{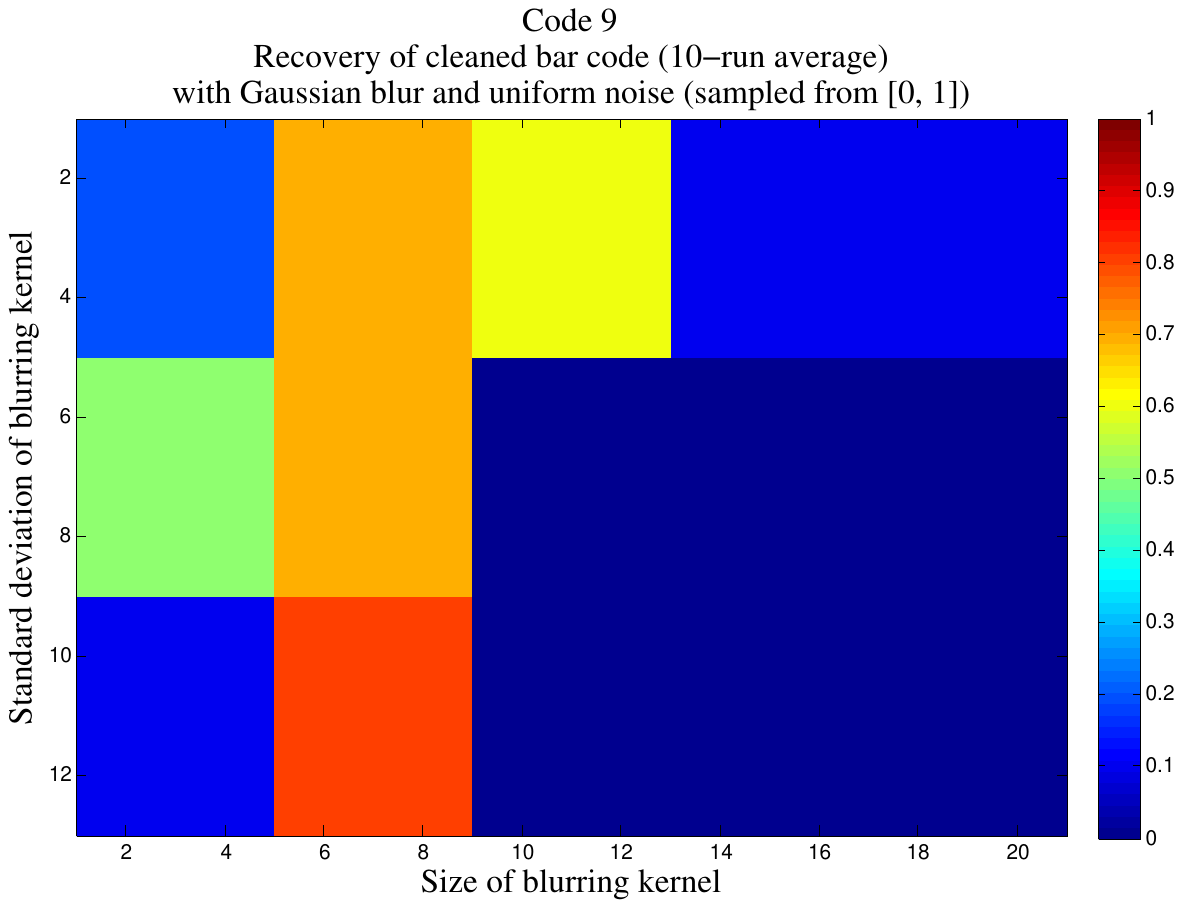}}

\end{center}
\caption{Readability for Code 9 from Figure~\ref{fig:code9}, for various sizes and standard deviations of the Gaussian blur kernel, with uniform noise (sampled from $[0,1]$)}
\label{fig:diaggaussgauss}
\end{figure}

 In Figure~\ref{fig:diaggaussgauss}  we see that unprocessed bar codes with Gaussian blur and uniform noise are completely unreadable by {\it ZBar} at the noise level we used. Our algorithm improves readability somewhat, but the results in this case are lacking. Furthermore, we notice a rather puzzling feature in Figure~\ref{fig:diaggaussgausscleaned} which is also consistently present in the results for the other bar codes: the readability results are better for blurring kernel size 7 than they are for size 3. We will address these issues in Section~\ref{sec:3methods}.

\begin{figure}[h]
\begin{center}

\subfloat{\label{fig:diagmotiongaussunprocessed}
\includegraphics[trim={0.5cm 6.9cm 0.7cm 6.8cm}, clip, width=0.48\columnwidth]{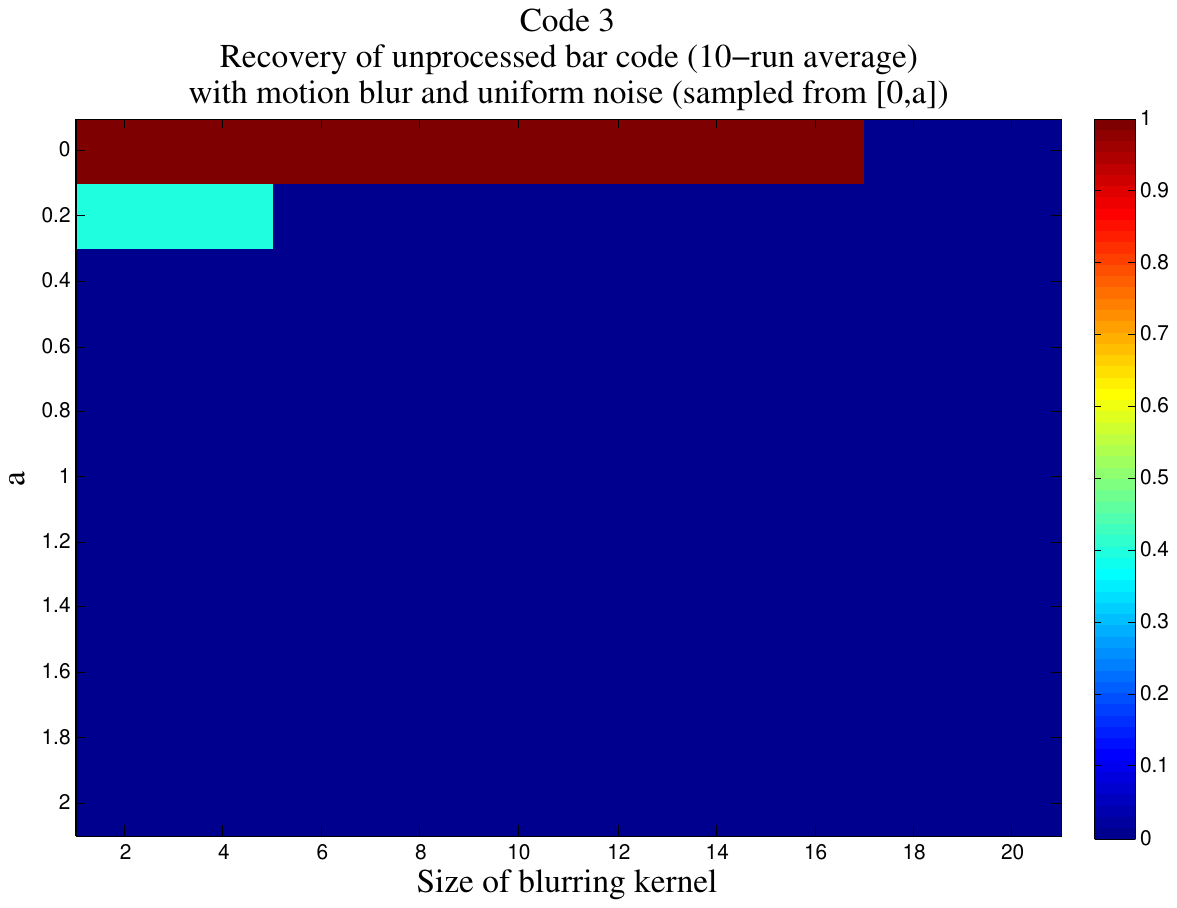}}
\subfloat{\label{fig:diagmotiongausscleaned}
\includegraphics[trim={0.5cm 6.9cm 0.7cm 6.8cm}, clip, width=0.48\columnwidth]{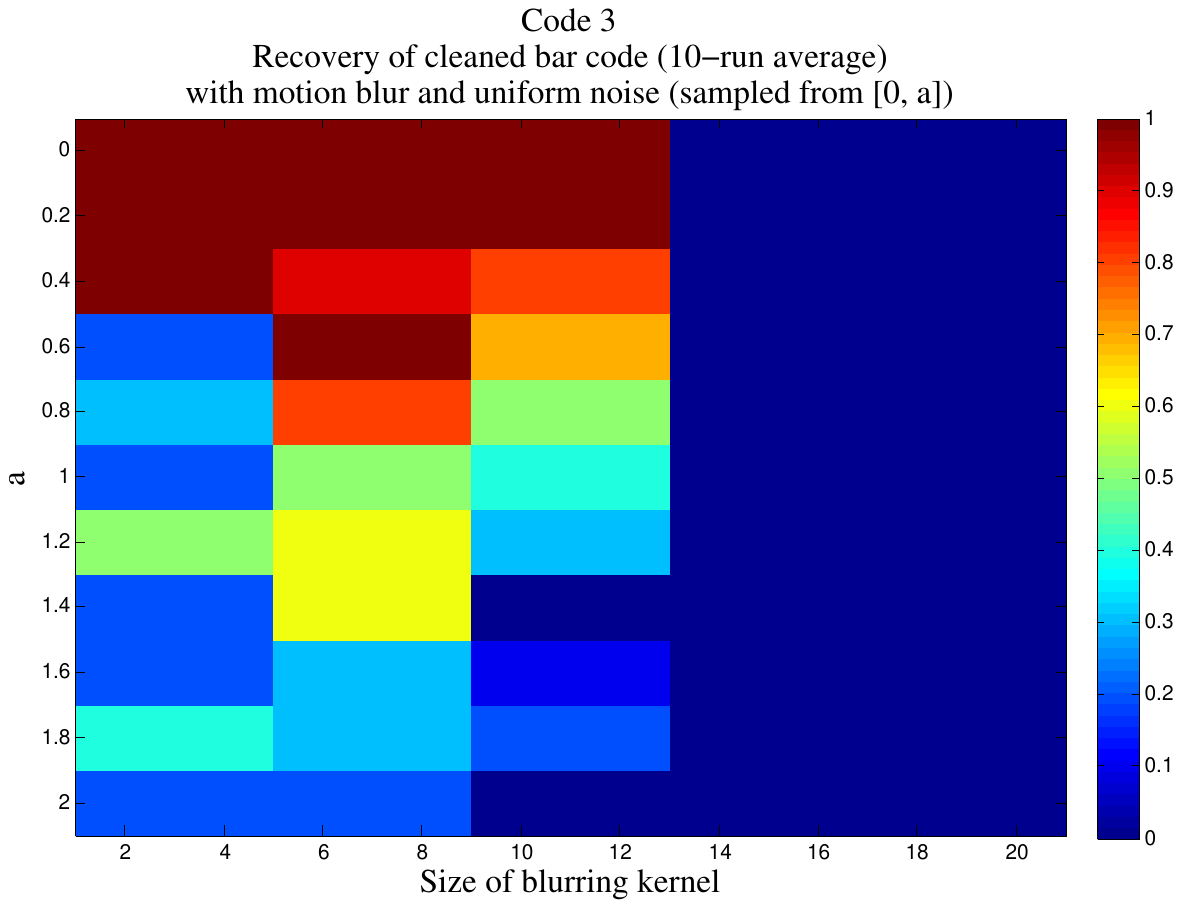}}
\end{center}
\caption{Readability of Code 3, for various blurring lengths and various lengths of the uniform noise's sampling interval $[0,a]$}
\label{fig:diagmotiongauss}
\end{figure}

In the presence of any uniform noise, the unprocessed motion blurred bar code is mostly unreadable (cf. Figure~\ref{fig:diagmotiongauss}). Our algorithm consistently makes low noise codes readable, while also leading to readability for nontrivial percentages of high noise level codes. We also notice a similar, although perhaps less emphatic, trend as in the Gaussian blur case (which again is present for all codes) for the results at blurring kernel size 7 to be better than those at size 3.
Once more we refer to Section~\ref{sec:3methods} for further discussion of this issue.

\subsubsection{Salt and pepper noise}
\begin{figure}[h]
\begin{center}

\subfloat{\label{fig:diaggausssandpunprocessed}
\includegraphics[trim={2.1cm 6.9cm 2cm 8.2cm}, clip, width=0.48\columnwidth]{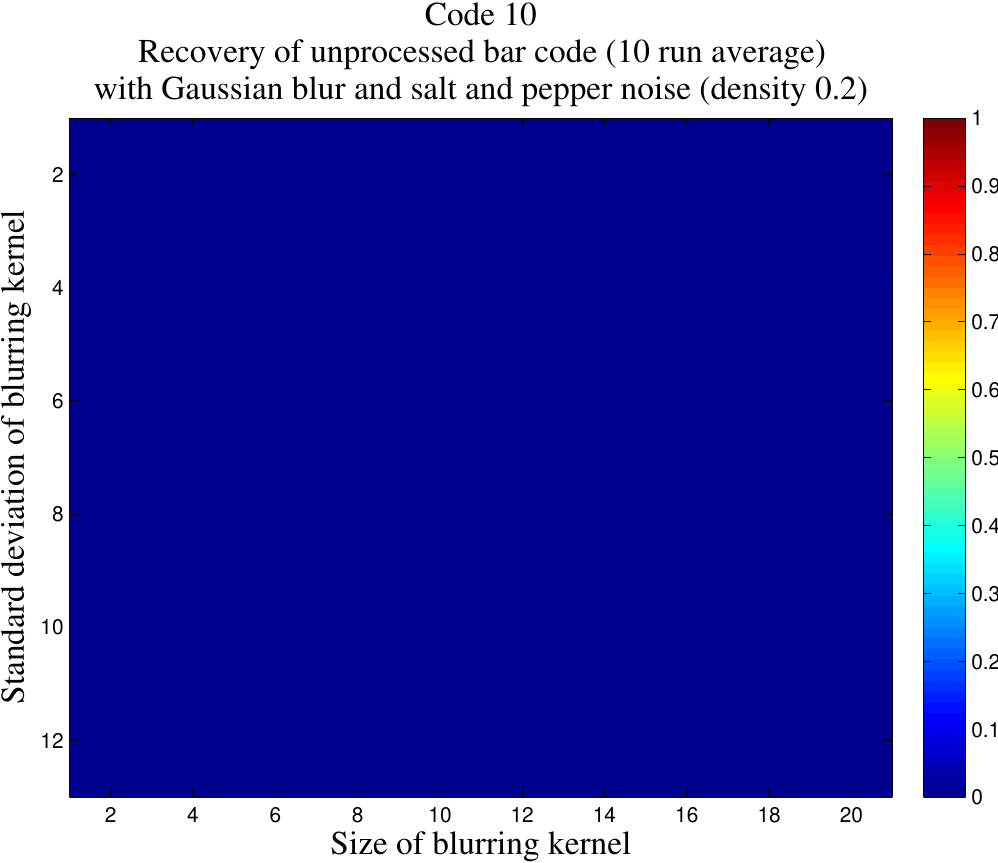}}
\subfloat{\label{fig:diaggausssandpcleaned}
\includegraphics[trim={2.1cm 6.9cm 2cm 8.2cm}, clip, width=0.48\columnwidth]{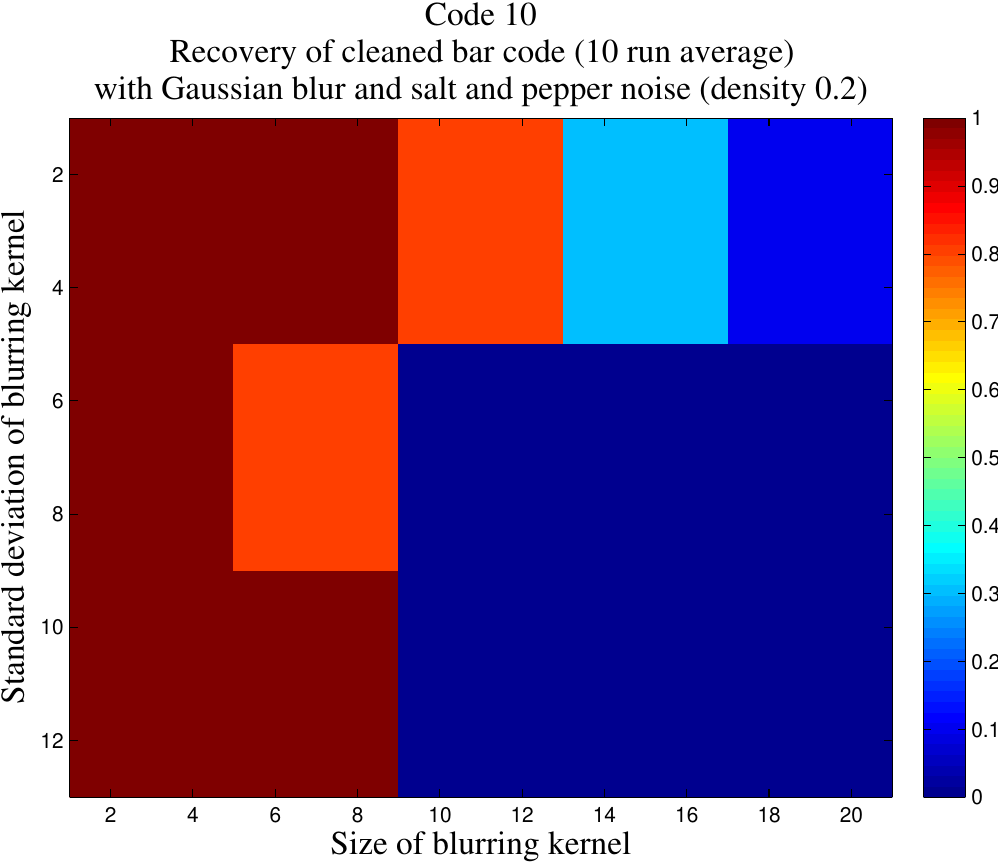}}

\end{center}
\caption{Readability of Code 10 from Figure~\ref{fig:code10}, for various sizes and standard deviations of the Gaussian blur kernel, with salt and pepper noise (density 0.2)}
\label{fig:diaggausssandp}
\end{figure}

\begin{figure}[h]
\begin{center}

\subfloat{\label{fig:diagmotionsandpunprocessed}
\includegraphics[trim={1.9cm 6.9cm 2cm 8.2cm}, clip, width=0.48\columnwidth]{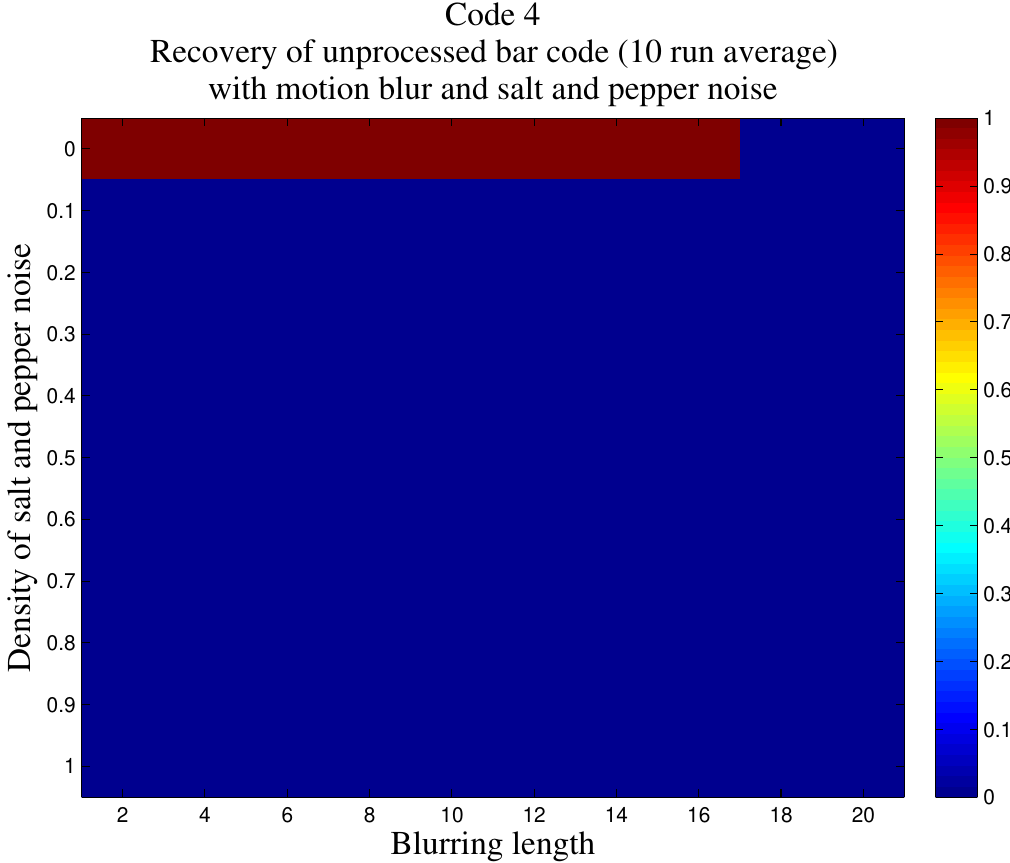}}
\subfloat{\label{fig:diagmotionsandpcleaned}
\includegraphics[trim={1.9cm 6.9cm 2cm 8.2cm}, clip, width=0.48\columnwidth]{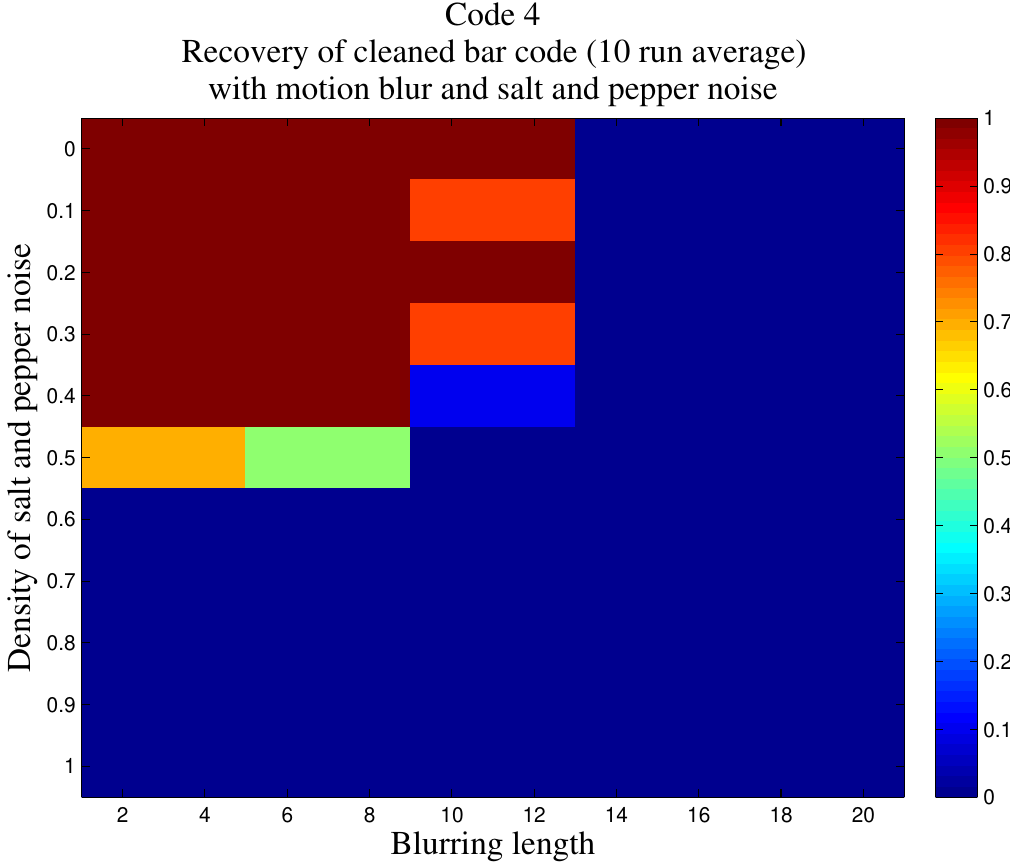}}

\end{center}
\caption{Readability of Code 4, for various blurring lengths and various values of the salt and pepper noise's density}
\label{fig:diagmotionsandp}
\end{figure}

Our algorithm performs consistently well for a range of blur and noise parameters, both for Gaussian blur and motion blur (cf. Figures~\ref{fig:diaggausssandp} and~\ref{fig:diagmotionsandp}). The uncleaned codes are completely unreadable in the presence of even the slightest noise.

\subsubsection{Speckle noise}

\begin{figure}[h]
\begin{center}

\subfloat{\label{fig:diaggaussspeckleunprocessed}
\includegraphics[trim={2.1cm 6.9cm 2cm 8.2cm}, clip, width=0.48\columnwidth]{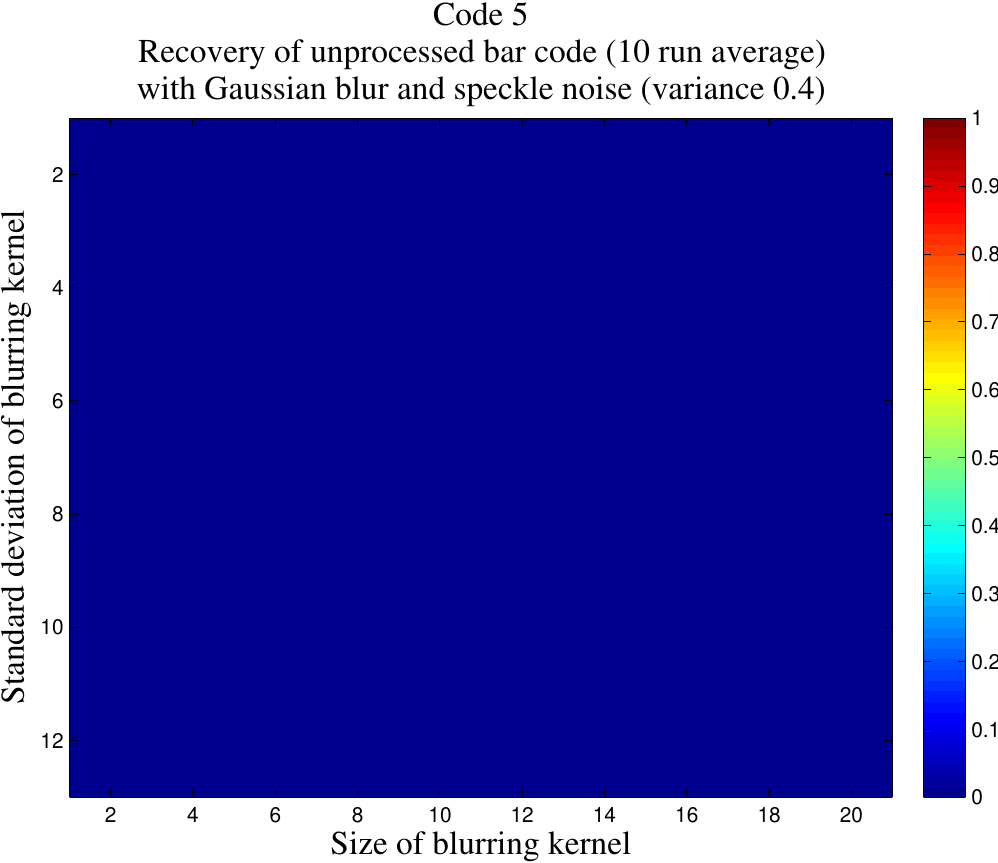}}
\subfloat{\label{fig:diaggaussspecklecleaned}
\includegraphics[trim={2.1cm 6.9cm 2cm 8.2cm}, clip, width=0.48\columnwidth]{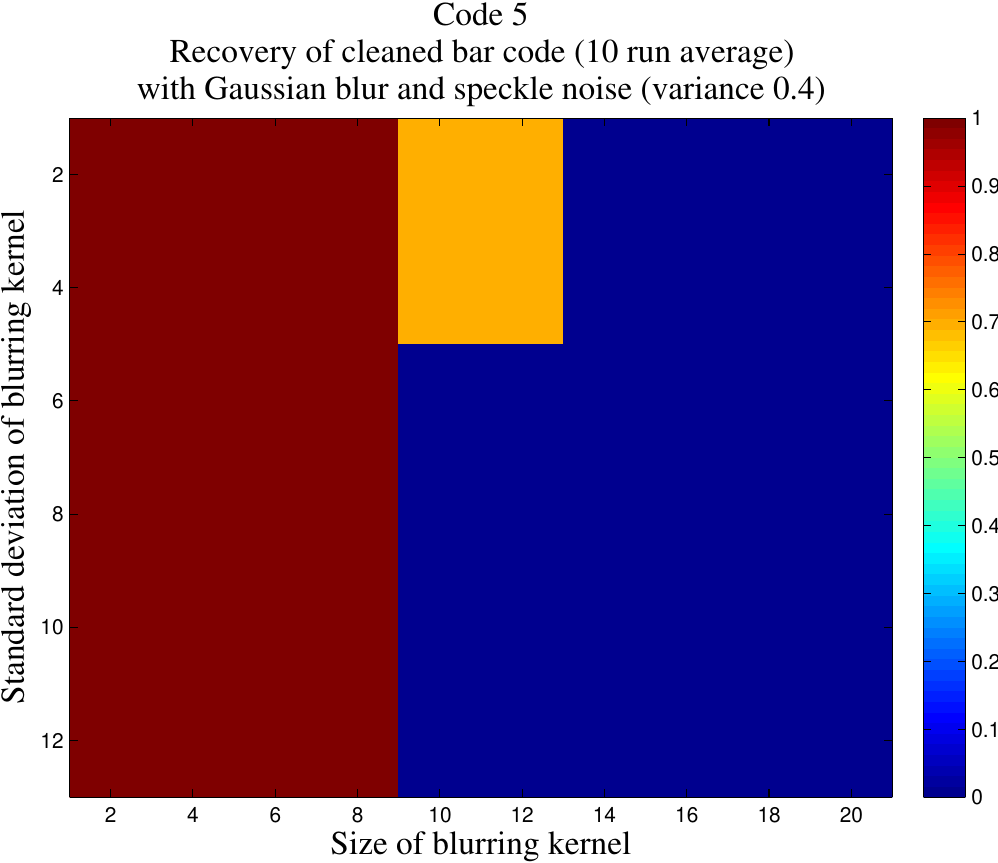}}

\end{center}
\caption{Readability of Code 5, for various sizes and standard deviations of the Gaussian blur kernel, with speckle noise (variance 0.4)}
\label{fig:diaggaussspeckle}
\end{figure}

\begin{figure}[h]
\begin{center}

\subfloat{\label{fig:diagmotionspeckleunprocessed}
\includegraphics[trim={1.9cm 6.9cm 2cm 8.2cm}, clip, width=0.48\columnwidth]{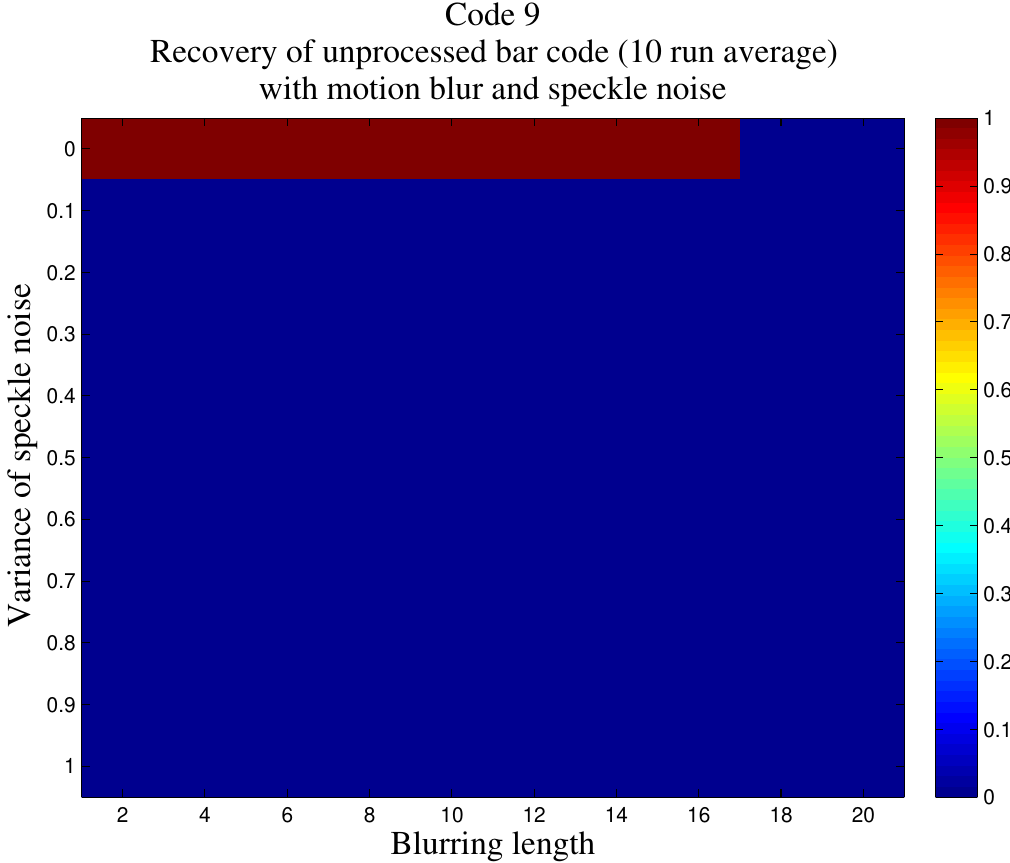}}
\subfloat{\label{fig:diagmotionspecklecleaned}
\includegraphics[trim={1.9cm 6.9cm 2cm 8.2cm}, clip, width=0.48\columnwidth]{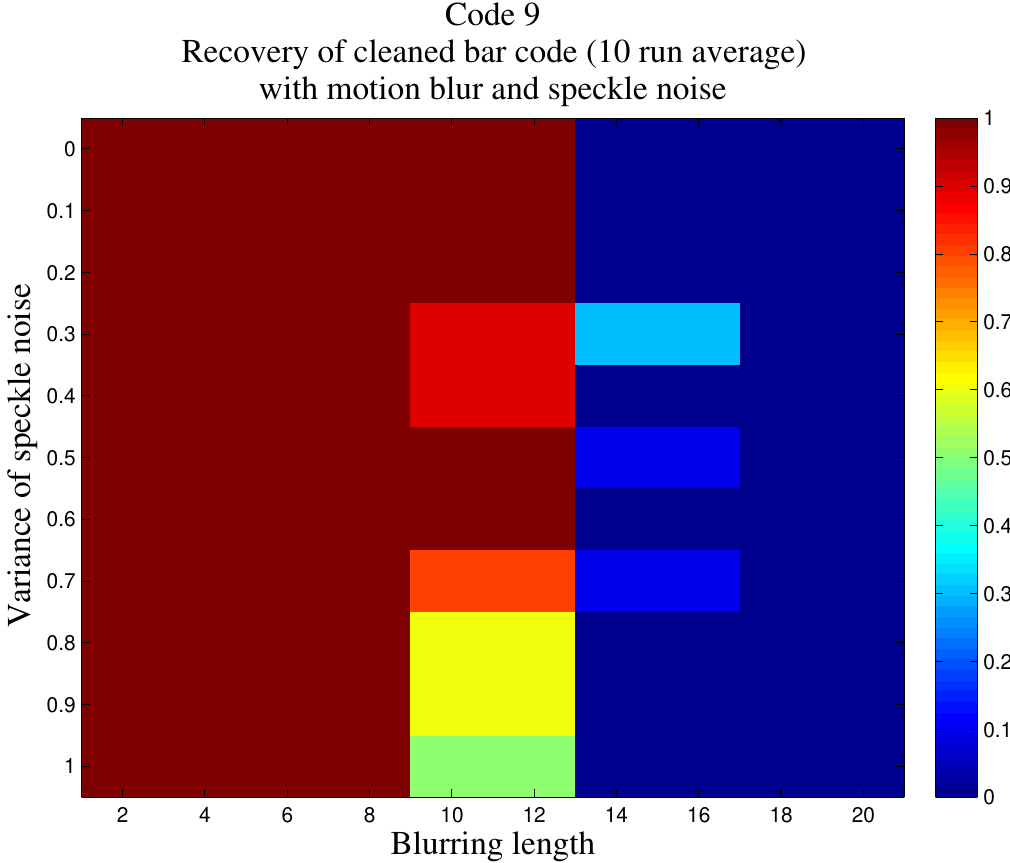}}

\end{center}
\caption{Readability of Code 9 from Figure~\ref{fig:code9}, for various blurring lengths and various values of the speckle noise's variance}
\label{fig:diagmotionspeckle}
\end{figure}

Results here are quite similar to those for salt and pepper noise (cf. Figures~\ref{fig:diaggaussspeckle} and~\ref{fig:diagmotionspeckle}). However, we note that the improvement in the readability of the motion blurred codes is
better than we have seen for any of the other types of noise above.

\subsection{Different choices for $\lambda_1$}\label{sec:varyinglambda1}

\begin{figure}[h]
\begin{center}
\subfloat{\label{fig:G1}
\includegraphics[trim={3.8cm 9.5cm 4.3cm 9.5cm}, clip, width=0.48\columnwidth]{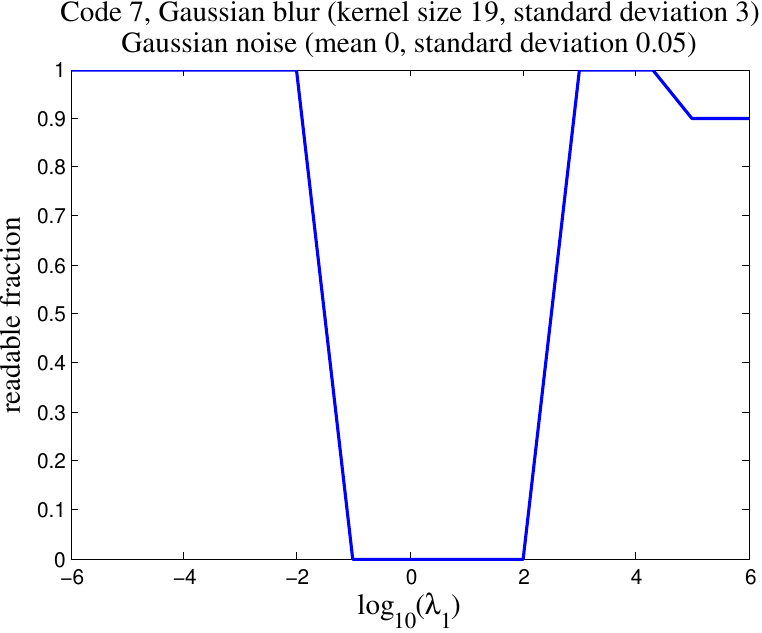}}
%\subfloat{\label{fig:G2}
%\includegraphics[trim={3.8cm 9.5cm 4.3cm 9.5cm}, clip, width=0.5\columnwidth]{varying_lambda1_Gaussblursdiam19Gaussblursd7Gaussnoisesd_pt05_Code7_retitled.pdf}}\\
\subfloat{\label{fig:g3}
\includegraphics[trim={3.8cm 9.5cm 4.3cm 9.5cm}, clip, width=0.48\columnwidth]{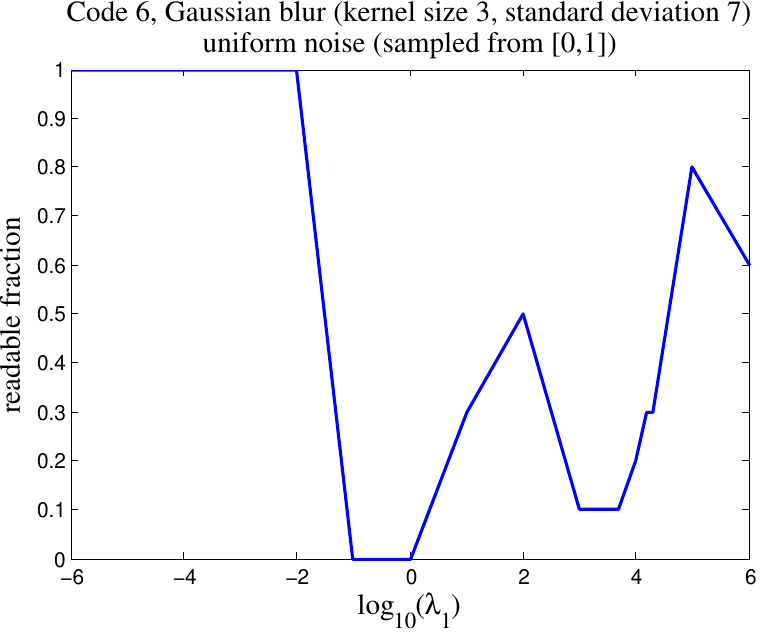}}\\
\subfloat{\label{fig:G4}
\includegraphics[trim={3.8cm 9.5cm 4.3cm 9.5cm}, clip, width=0.48\columnwidth]{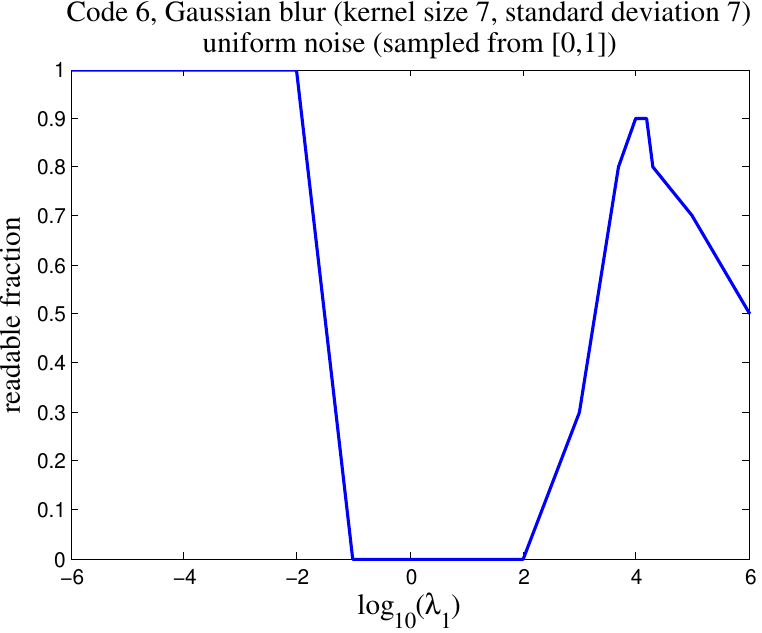}}
%\subfloat{\label{fig:G5}
%\includegraphics[trim={3.8cm 9.5cm 4.3cm 9.5cm}, clip, width=0.5\columnwidth]{varying_lambda1_Gausblursdiam19Gaussblursd3unifnoise0-1_Code9_retitled.pdf}}
%\subfloat{\label{fig:G6}
%\includegraphics[trim={3.8cm 9.5cm 4.3cm 9.5cm}, clip, width=0.5\columnwidth]{varying_lambda1_Gausblursdiam19Gaussblursd7unifnoise0-1_Code9_retitled.pdf}}\\
%\subfloat{\label{fig:G7}
%\includegraphics[trim={3.8cm 9.5cm 4.3cm 9.5cm}, clip, width=0.5\columnwidth]{varying_lambda1_Gausblursdiam3Gaussblursd3saltpepper_dens_pt2_Code10_retitled.pdf}}
%\subfloat{\label{fig:G8}
%\includegraphics[trim={3.8cm 9.5cm 4.3cm 9.5cm}, clip, width=0.5\columnwidth]{varying_lambda1_Gausblursdiam7Gaussblursd11saltpepper_dens_pt2_Code10_retitled.pdf}}\\
%\subfloat{\label{fig:G9}
%\includegraphics[trim={3.8cm 9.5cm 4.3cm 9.5cm}, clip, width=0.5\columnwidth]{varying_lambda1_Gausblursdiam15Gaussblursd7saltpepper_dens_pt2_Code10_retitled.pdf}}
\subfloat{\label{fig:G10}
\includegraphics[trim={3.8cm 9.5cm 4.3cm 9.5cm}, clip, width=0.48\columnwidth]{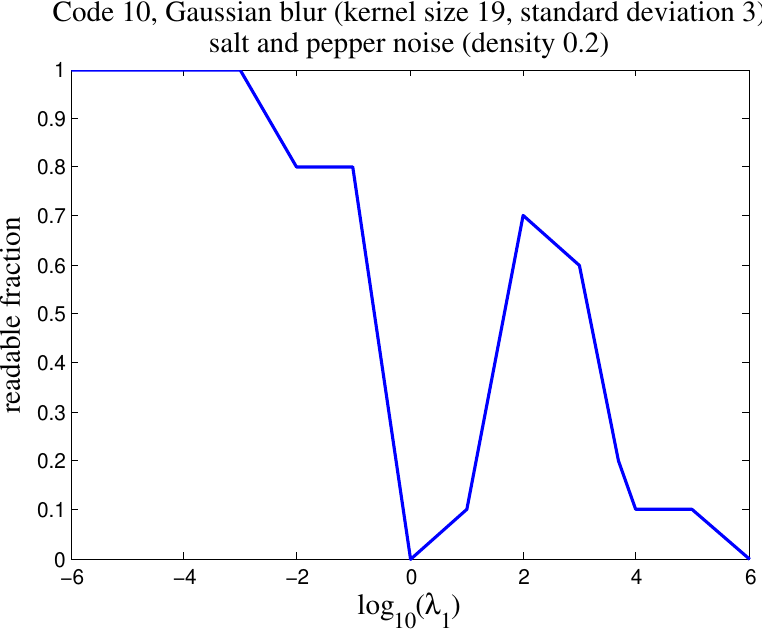}}\\
\subfloat{\label{fig:G11}
\includegraphics[trim={3.8cm 9.5cm 4.3cm 9.5cm}, clip, width=0.48\columnwidth]{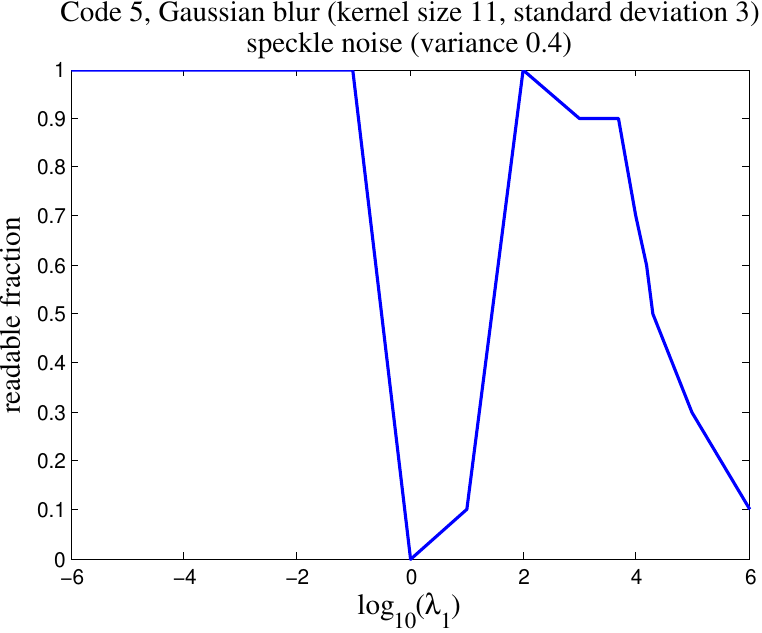}}
\subfloat{\label{fig:G12}
\includegraphics[trim={3.8cm 9.5cm 4.3cm 9.5cm}, clip, width=0.48\columnwidth]{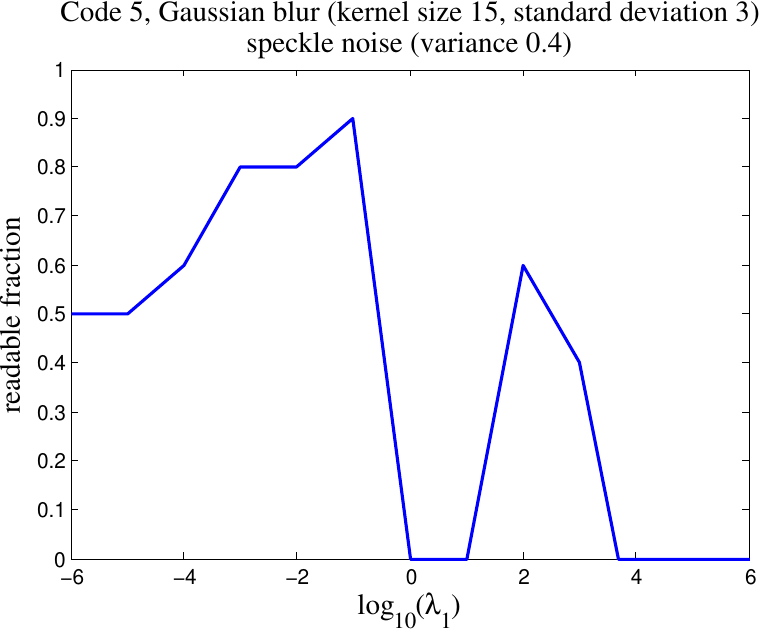}}\\
\caption{Readability (as fraction of 10 runs with different noise realizations) of codes with Gaussian blur and different types of noise, as function of $\log_{10}(\lambda_1)$}\label{fig:lambda1graphsGaussian}
\end{center}
\end{figure}

\begin{figure}[h]
\begin{center}
\subfloat{\label{fig:1}
\includegraphics[trim={3.8cm 9.5cm 4.3cm 9.5cm}, clip, width=0.48\columnwidth]{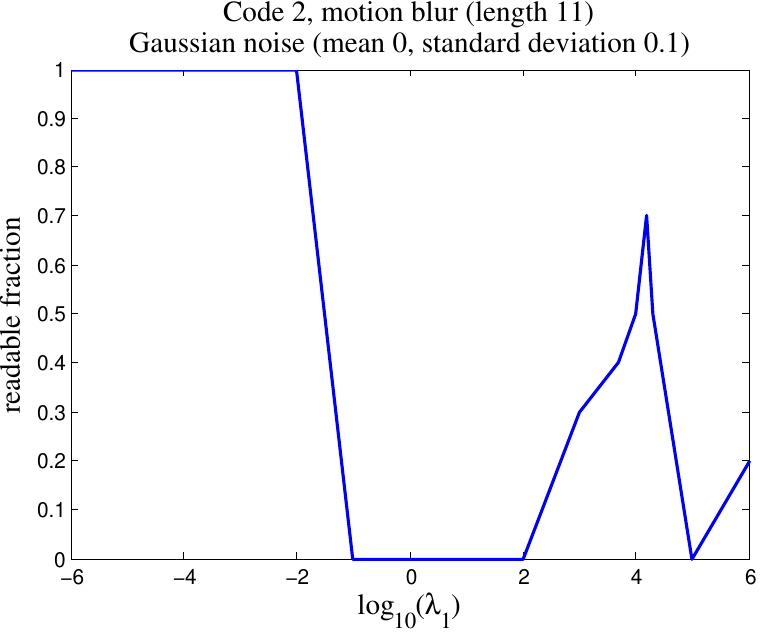}}
%\subfloat{\label{fig:2}
%\includegraphics[trim={3.8cm 9.5cm 4.3cm 9.5cm}, clip, width=0.5\columnwidth]{varying_lambda1_Motionblurlen15Motionblurangle30_Gaussnoisesd_pt05_Code2_retitled.pdf}}\\
%\subfloat{\label{fig:3}
%\includegraphics[trim={3.8cm 9.5cm 4.3cm 9.5cm}, clip, width=0.5\columnwidth]{varying_lambda1_Motionblurlen15Motionblurangle30_unifnoise0-1_Code3_retitled.pdf}}
\subfloat{\label{fig:4}
\includegraphics[trim={3.8cm 9.5cm 4.3cm 9.5cm}, clip, width=0.48\columnwidth]{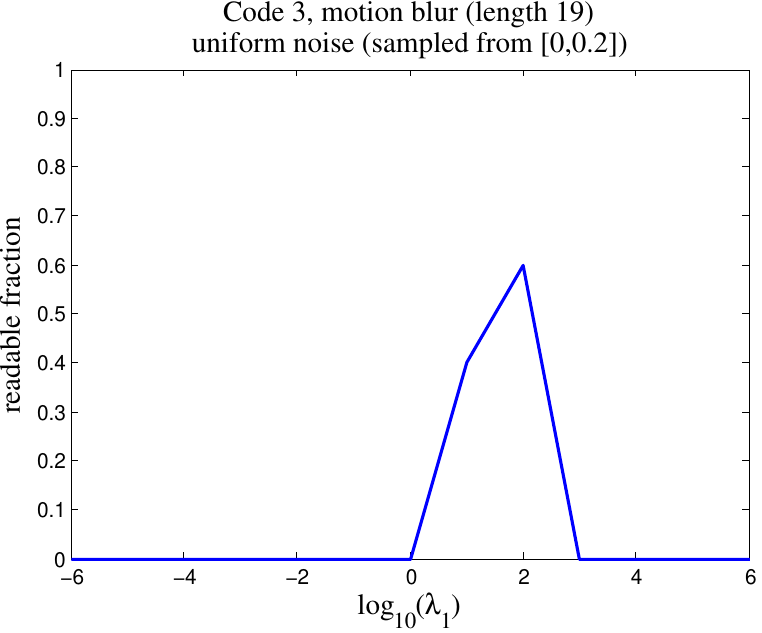}}\\
%\subfloat{\label{fig:5}
%\includegraphics[trim={3.8cm 9.5cm 4.3cm 9.5cm}, clip, width=0.5\columnwidth]{varying_lambda1_Motionblurlen3Motionblurangle30_saltpepper_dens_pt2_Code4_retitled.pdf}}
%\subfloat{\label{fig:6}
%\includegraphics[trim={3.8cm 9.5cm 4.3cm 9.5cm}, clip, width=0.5\columnwidth]{varying_lambda1_Motionblurlen7Motionblurangle30_saltpepper_dens_pt3_Code4_retitled.pdf}}\\
\subfloat{\label{fig:7}
\includegraphics[trim={3.8cm 9.5cm 4.3cm 9.5cm}, clip, width=0.48\columnwidth]{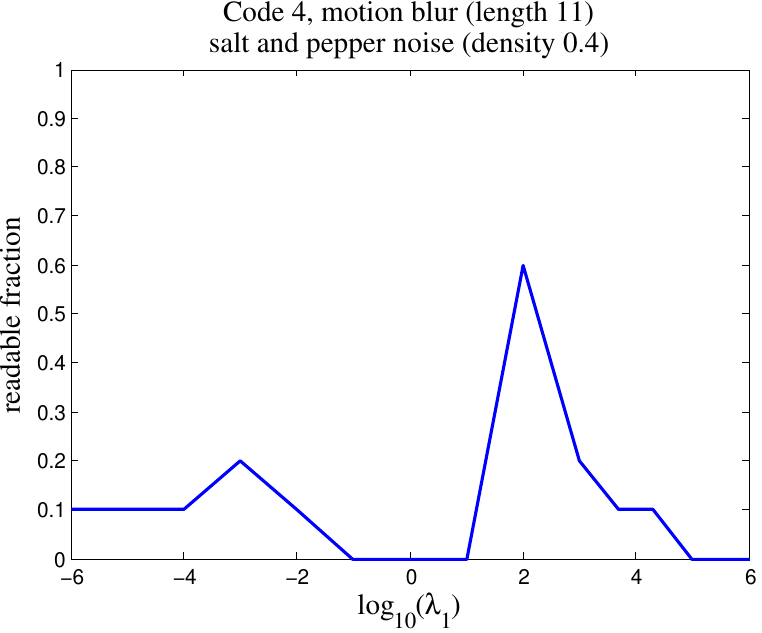}}
\subfloat{\label{fig:8}
\includegraphics[trim={3.8cm 9.5cm 4.3cm 9.5cm}, clip, width=0.48\columnwidth]{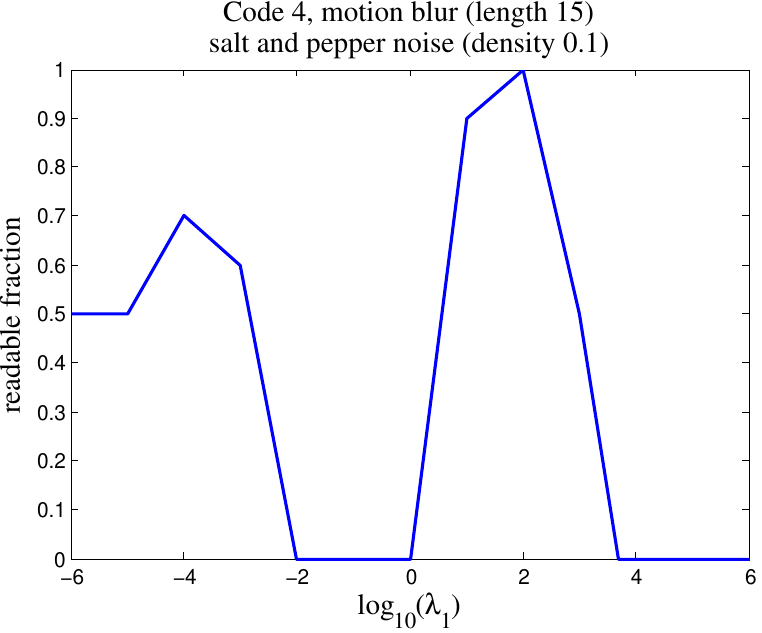}}\\
\subfloat{\label{fig:9}
\includegraphics[trim={3.8cm 9.5cm 4.3cm 9.5cm}, clip, width=0.48\columnwidth]{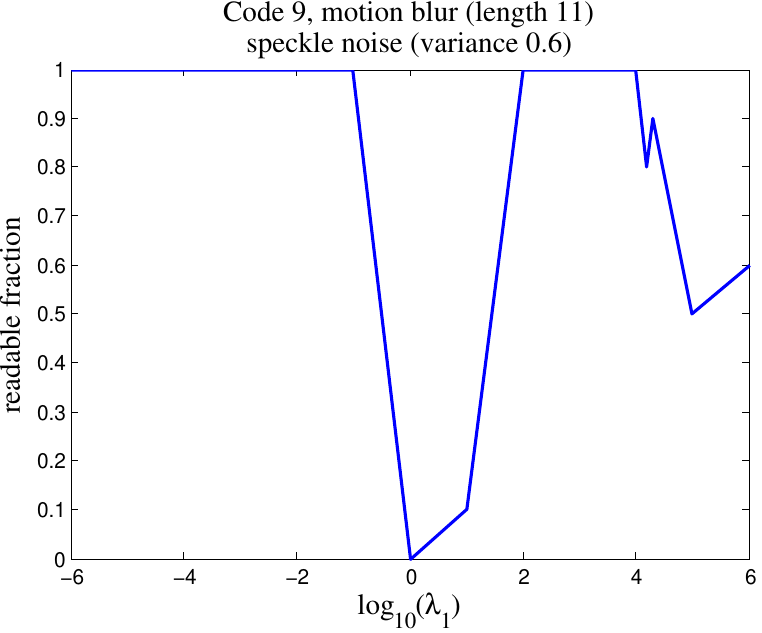}}
\subfloat{\label{fig:10}
\includegraphics[trim={3.8cm 9.5cm 4.3cm 9.5cm}, clip, width=0.48\columnwidth]{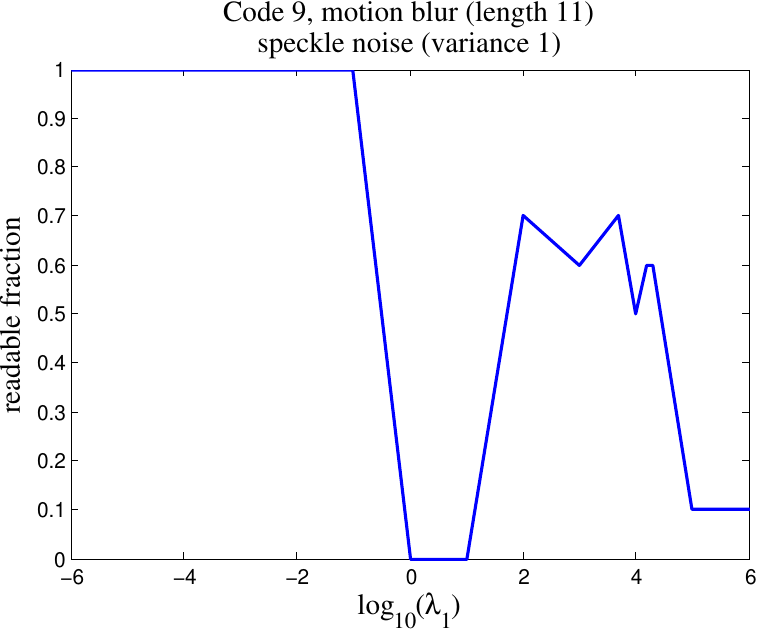}}\\
%\subfloat{\label{fig:11}
%\includegraphics[trim={3.8cm 9.5cm 4.3cm 9.5cm}, clip, width=0.5\columnwidth]{varying_lambda1_Motionblurlen15Motionblurangle30_speckle_var_pt7_Code9_retitled.pdf}}
\caption{Readability (as fraction of 10 runs with different noise realizations) of codes with motion blur and different types of noise, as function of $\log_{10}(\lambda_1)$}\label{fig:lambda1graphsmotion}
\end{center}
\end{figure}

The choice of the parameter $\lambda_1$ in the PSF estimation step (described in Section~\ref{sec:PSFestimation}) is an important one. In all the results we have discussed so far, we used $\lambda_1=10000$. This choice was based on some initial trial and error runs with various values for $\lambda_1$. Here we look more closely at the influence of the choice of $\lambda_1$,  by running the algorithm  on a selection of codes for values of $\lambda_1$ ranging\footnote{Specifically,  $\lambda_1\in \{10^{-6}, 10^{-5}, \ldots, 10^2, 10^3, \break 5 \times 10^3, 10^4, 1.5 \times 10^4, 10^5, 10^6\}$.} from $10^{-6}$ to $10^6$. The results are depicted in  Figures~\ref{fig:lambda1graphsGaussian} (Gaussian blur) and~\ref{fig:lambda1graphsmotion} (motion blur).

We see that the specific dependence of the performance on $\lambda_1$ differs per code, but there are a few general trends we can discern. If the blurring is not too strong, then $0<\lambda_1<0.1$ performs well  for both Gaussian and motion blur. This suggests that in these cases a constant PSF might perform equally well or even better than the PSF estimation of step (ii). We further investigate this in Section~\ref{sec:3methods}.

In most cases there is a clear dip in performance around $\lambda_1=1$, implying that ``the middle road" between constant PSF ($\lambda_1 =0$) and ``high fidelity" (large $\lambda_1$) reconstructed PSF is never the preferred choice. In the presence of heavy blurring, values of $\lambda_1$ between $10^2$ and $10^4$ do typically perform well, especially in the case of motion blur. This is also supported by  the fact that our previous choice of $\lambda_1=10000$  produced good results.

\subsection{Three different algorithms}\label{sec:3methods}

Three issues from our discussion of the results remain to be addressed: (a) in Section~\ref{sec:resultsalg} we saw noiseless codes with heavy motion blur (length 15) are readable in unprocessed form, but not after being cleaned (with our standard algorithm, i.e. $\lambda_1=10000$); (b) cleaned codes with uniform noise are less readable with very light blurring than with slightly heavier blurring (especially for Gaussian blur) as discussed in Section~\ref{sec:uniformnoise}; (c) the results from Section~\ref{sec:varyinglambda1} imply that for light blurring a constant PSF could perform as well or better than the PSF reconstructed by our PSF estimation step.

To investigate these issues, we run three different variations of our algorithm on a selection of bar codes. ``Denoising only'' (D) entails performing only steps (i) and (iv) (Sections~\ref{sec:denoisingviaTVflow} and~\ref{sec:thresholdingstep}) from our full algorithm. ``Uniform PSF'' (UPSF) keeps steps (i), (iii), and (iv), but changes the PSF estimation step (ii). Instead of solving a variational problem for $\phi_*$, we let $\phi_*$ be constant and then choose the optimal size for the kernel by minimizing $\|\phi_i*z - u_1\|_{L^2(C_1)}^2$ as in Section~\ref{sec:PSFestimation}. Heuristically this corresponds to choosing $\lambda_1=0$ in the original step (ii). Finally ``Full PSF'' (FPSF) refers to our original algorithm, which we run for the same range of $\lambda_1$ values as described in Section~\ref{sec:varyinglambda1}.

\begin{figure}[h]
\begin{center}

\subfloat{\label{fig:denoisingonlyGauss}
\includegraphics[trim={1.3cm 7.6cm 0.8cm 7cm}, clip, width=0.48\columnwidth]{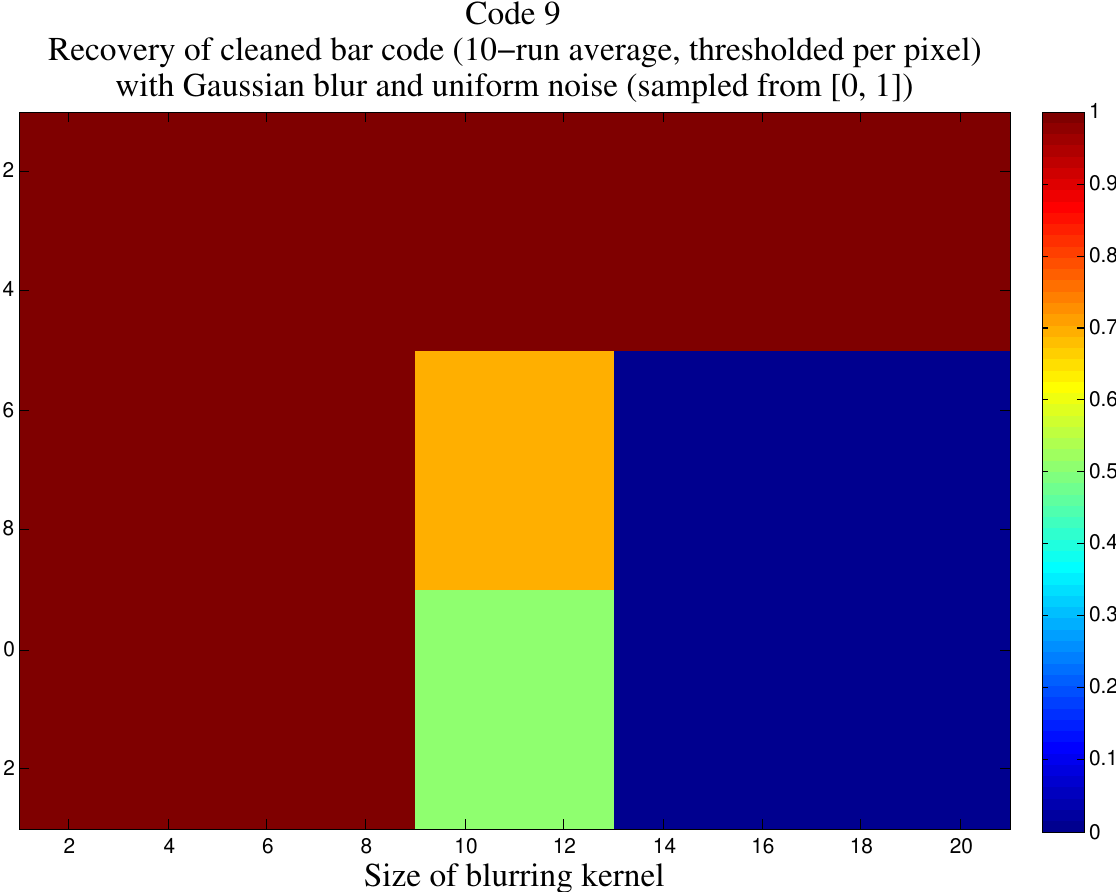}}
\subfloat{\label{fig:denoisingonlymotion}
\includegraphics[trim={1.3cm 7.6cm 0.8cm 7cm}, clip, width=0.48\columnwidth]{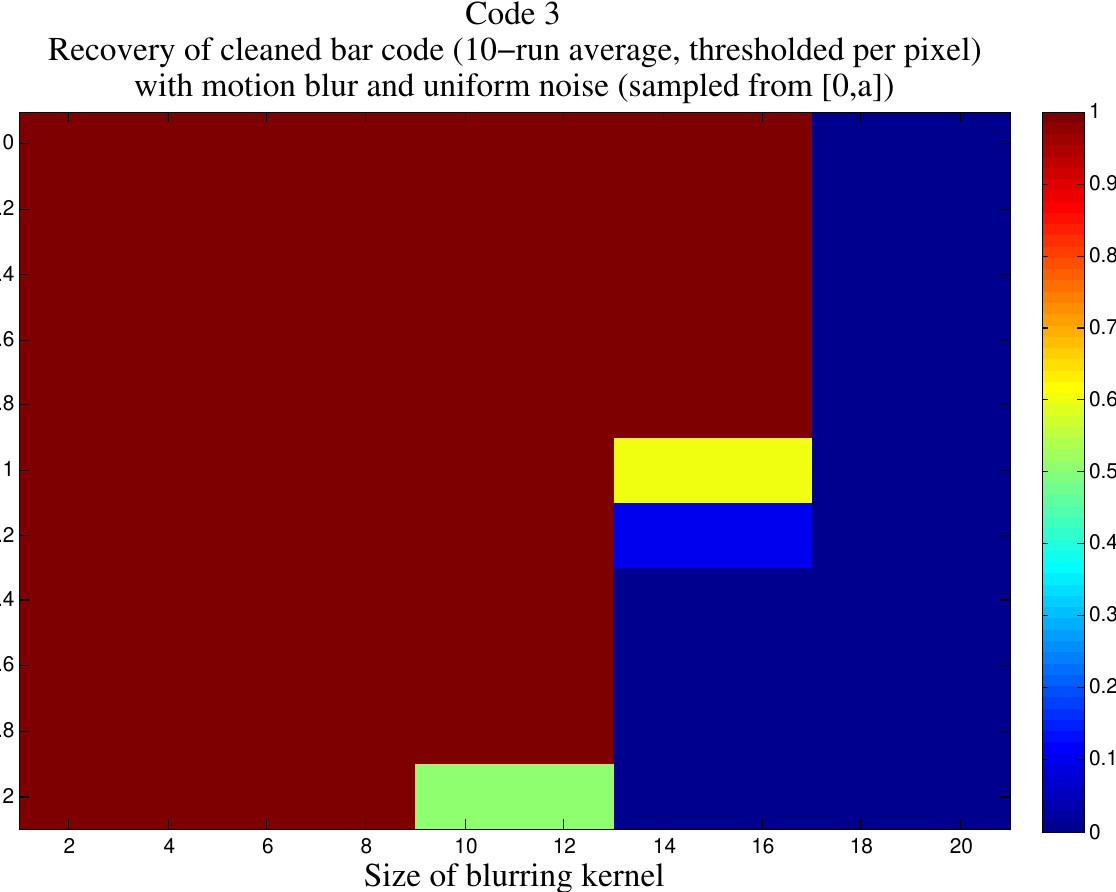}}

\end{center}
\caption{Representative results for the ``Denoising only'' algorithm in the presence of uniform noise}
\label{fig:denoisingonly}
\end{figure}

\begin{figure}[h]
\begin{center}

\subfloat{\label{fig:UPSFGauss}
\includegraphics[trim={1.3cm 7.6cm 0.8cm 7cm}, clip, width=0.48\columnwidth]{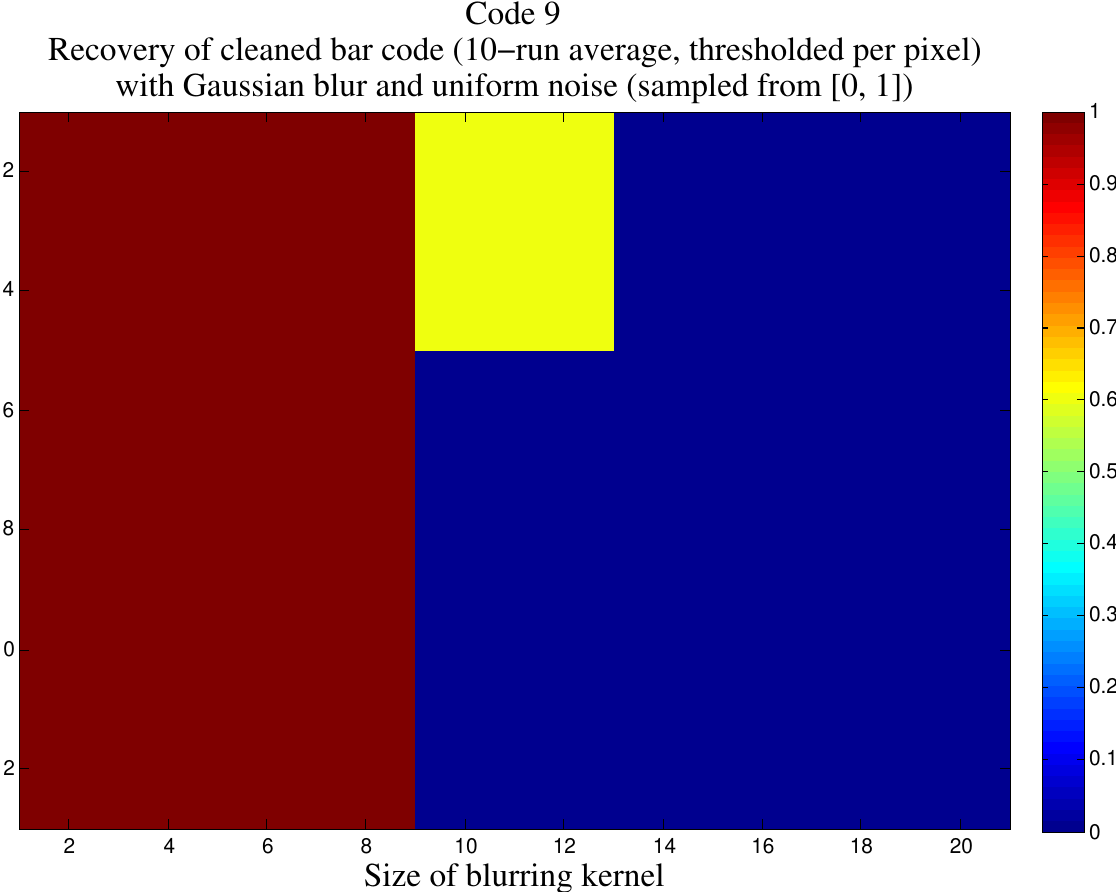}}
\subfloat{\label{fig:UPSFmotion}
\includegraphics[trim={1.3cm 7.6cm 0.8cm 7cm}, clip, width=0.48\columnwidth]{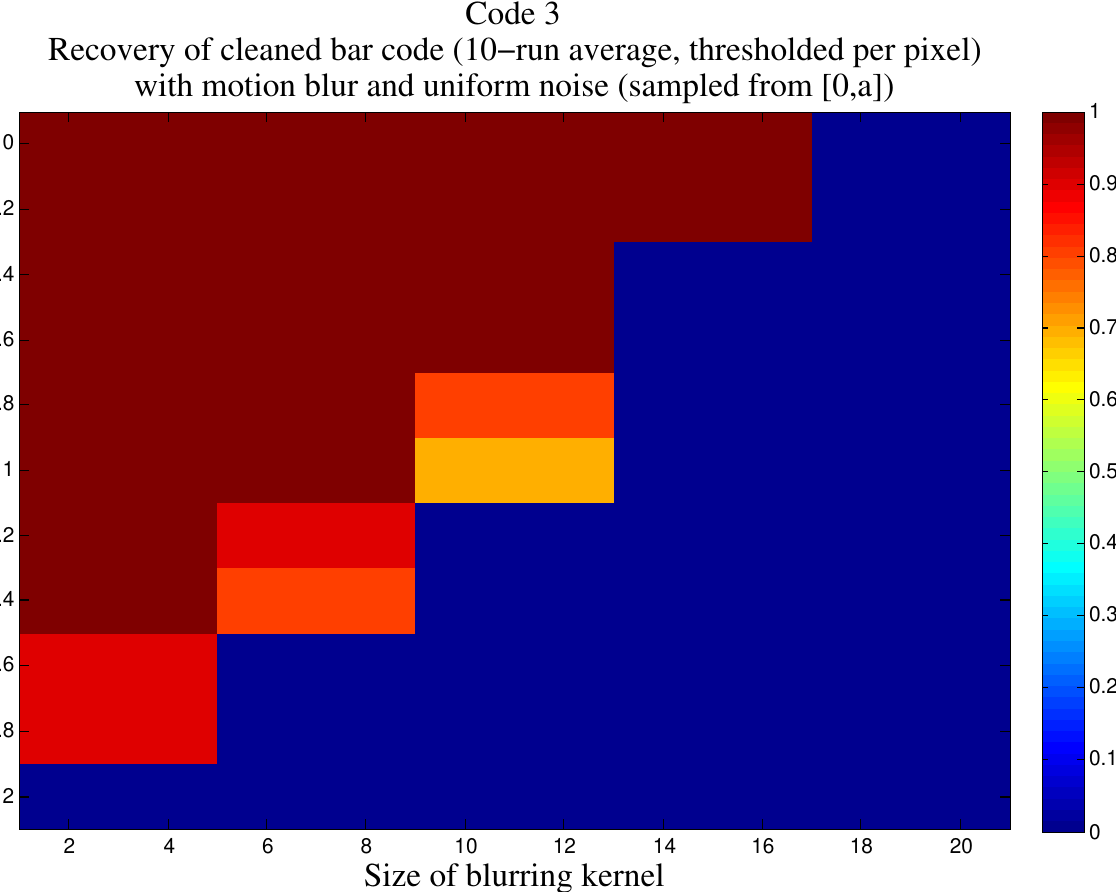}}

\end{center}
\caption{Representative results for the ``Uniform PSF'' algorithm in the presence of uniform noise}
\label{fig:UPSF}
\end{figure}

Figures~\ref{fig:denoisingonly} and~\ref{fig:UPSF} show representative results for (D) and (UPSF), respectively, on codes with uniform noise. Comparing with Figures~\ref{fig:diaggaussgauss} and~\ref{fig:diagmotiongauss}, we see that both issue (a) and issue (b) disappear in these cases and that (D) performs even better than (UPSF), suggesting that the PSF estimation step is not beneficial in the presence of uniform noise.
\begin{table}\scriptsize
\centering
{
\begin{tabular}{|>{\centering}p{0.1cm}|>{\centering}p{0.8cm}|>{\centering}p{0.8cm}|>{\centering}p{0.1cm}|>{\centering}p{0.1cm}|>{\centering}p{0.4cm}|>{\centering}p{0.4cm}|>{\centering}p{1.1cm}|}
\hline
%\textbf{Code}&\textbf{Blur}&\textbf{Noise}&\textbf{Unprocessed}&\textbf{Denoising only}&\textbf{Uniform PSF}&\textbf{Full PSF}\\\hline
\textbf{C}&\textbf{B}&\textbf{N}&\textbf{U}&\textbf{D}&\textbf{UPSF}&\textbf{FPSF}&\textbf{$\lambda_1$}\tabularnewline\hhline{|=|=|=|=|=|=|=|=|}
7 & $G(19,3)$ & $G(0.5)$ & 0 & 10 & 10 & 10 & S, $10^3$ $5 \cdot 10^3$ $10^4$ $1.5 \cdot 10^4$\tabularnewline\hline
7 & $G(19,7)$ & $G(0.5)$ & 0 & 0 & 0 & 0 & A\tabularnewline\hline
9 & $G(19,3)$ & $U(1)$ & 0 & 10 & 0 & 1 & $10^3$ $5 \cdot 10^3$ $10^4$ $1.5 \cdot 10^4$\tabularnewline\hline
9 & $G(19,7)$ & $U(1)$ & 0 & 0 & 0 & 0 & A\tabularnewline\hline
10 & $G(15,7)$ & $Sa(0.2)$ & 0 & 0 & 0 & 0 & A\tabularnewline\hline
10 & $G(19,3)$ & $Sa(0.2)$ & 0 & 0 & 10 & 10 & S\tabularnewline\hline
5 & $G(11,3)$ & $Sp(0.4)$ & 0 & 6 & 10 & 10 & S, $10^2$\tabularnewline\hline
5 & $G(15,3)$ & $Sp(0.4)$ & 0 & 0 & 5 & 9 & S\tabularnewline\hhline{|=|=|=|=|=|=|=|=|}
2 & $M(11)$ & $G(0.1)$ & 0 & 10 & 10 & 10 & S\tabularnewline\hline
2 & $M(15)$ & $G(0.05)$ & 0 & 10 & 9 & 10 & S, $10^2$\tabularnewline\hline
3 & $M(15)$ & $U(1)$ & 0 & 6 & 0 & 0 & A\tabularnewline\hline
3 & $M(19)$ & $U(0.2)$ & 0 & 0 & 0 & 6 & $10^2$\tabularnewline\hline
4 & $M(11)$ & $Sa(0.4)$ & 0 & 0 & 1 & 6 & $10^2$\tabularnewline\hline
4 & $M(15)$ & $Sa(0.1)$ & 0 & 6 & 5 & 10 & $10^2$\tabularnewline\hline
9 & $M(11)$ & $Sp(1)$ & 0 & 1 & 10 & 10 & S\tabularnewline\hline
9 & $M(15)$ & $Sp(0.7)$ & 0 & 0 & 0 & 1 & $10^2, 10^3$ $10^4$ $1.5 \cdot 10^4$\tabularnewline\hline
\end{tabular}
\vspace{0.3cm}
}
\caption{Readability of a sample of bar codes.  Column {\bf C}: the number of clean bar codes used from our ten code
 catalogue. Column {\bf B}: $G(m,s)$ (Gaussian blur with mean $m$ and standard deviation $s$), $M(l)$
 (motion blur of length $l$). Column {\bf N}: $G(s)$ (Gaussian noise with mean 0 and standard deviation $s$), $U(x)$ (uniform noise sampled from $[0,x]$), $Sa(d)$ (salt and pepper noise with density $d$), $Sp(v)$ (speckle noise with variance $v$).
 The columns {\bf U} (``Unprocessed''), {\bf D} (``Denoising only'') and {\bf UPSF} (``Uniform PSF'') show the  number of
 {\it ZBar}-readable codes out of ten realizations of the noise per code.
 The {\bf FPSF} (``Full PSF'') column lists the same, for the values of $\lambda_1$ (listed in the last column) that give the best readability. Here ``S'' indicates small $\lambda_1$, i.e., $0<\lambda_1\leq 0.1$,
 ``A'' indicates the results were the same for all values of $\lambda_1$  tried.}
\label{tab:differentcases}
\end{table}

Table~\ref{tab:differentcases} lists results for a selection of codes representing all blurring-noise combinations. We also refer back to the graphs in Figures~\ref{fig:lambda1graphsGaussian} and~\ref{fig:lambda1graphsmotion}, which correspond to some of the (FPSF) results in the table. We note that the performance of (UPSF) is comparable with that of (FPSF), if the latter has small $\lambda_1$.
% as is expected (see Section~\ref{sec:varyinglambda1}).
We also note that (FPSF) with $\lambda_1=100$ performs well in the presence of heavy motion blurring and in some cases clearly outperforms the other algorithms; this is most striking for salt and pepper noise and for low uniform noise (with very heavy blurring). This confirms there are cases where the PSF estimation step improves the readability of the output considerably.

\section{Final Comments}
We have presented, and tested with {\it Zbar}, a regularization based algorithm for blind deblurring and denoising of QR bar codes.
The strength of our method is that it is ansatz-free with respect to the structure of the PSF and the noise.  In particular, it can deal with motion blurring. Note that we have focused entirely on regularization based methods for their simplicity of implementation, modest memory requirements, and speed.
More intricate denoising techniques could certainly be employed, for example,
patch-based denoising techniques are currently state of the art (\cite{Dabov, Milanfar}). One of the most efficient is BM3D (block matching 3D) which is based upon the effective filtering in a 3D transform domain, built by a combination of similar patches detected and registered by a block-matching approach. While such methods may indeed  provide better  denoising results, they have two main drawbacks: they require a considerable amount of memory to store the 3D domain and  are time consuming.  QR barcodes are widely read by smartphones and hence we have chosen a denoising algorithm which is easy to implement, requires little memory, and is computationally efficient.

With respect to comparisons of our method with other known methods for deblurring, we make a few comments.
One of the benefits of working with QR bar codes is that we do not need to, or want to, assess our results with  the ``eye-ball norm", but rather with bar code reading software such as the open source {\it ZBar}.
While we know of no other method specifically designed for blind deblurring and denoising of QR codes, the closest method is that of \cite{Liuetal} for 2D matrix codes. That method is based upon a Gaussian ansatz for noise and blurring, but also exploits a finder pattern, which in that case is an L-shaped corner.  While the authors use a different reading software, {\it ClearImage} \cite{ClearImage}, it would be interesting to adapt their method to QR codes, and compare their results with ours for Gaussian blurring and noise. We have not included such a comparison in the present paper, because we do not have access to either the same hardware (printer, camera) or software used in \cite{Liuetal}. Since the results in that paper depend heavily on these factors, any comparison produced with different tools would be misleading.

\section*{Acknowledgment}

A significant part of the first and third authors' work was performed while they were at the Department of Mathematics at UCLA.
The work of RC was partially supported by an NSERC (Canada) Discovery Grant. We thank the  referees for their useful comments on an earlier version of this paper which led to substantial improvements.

\end{document}